\documentclass[11pt]{article}

\usepackage[preprint]{acl}

\usepackage{times}
\usepackage{latexsym}

\usepackage[T1]{fontenc}

\usepackage[utf8]{inputenc}

\usepackage{microtype}

\usepackage{inconsolata}

\usepackage{graphicx}

\usepackage{amsmath} 

\usepackage{algorithm} 

\usepackage{algorithmic}

\usepackage{amssymb}

\usepackage{booktabs}   
\usepackage{tabularx}   
\usepackage{geometry}   
\usepackage{makecell}   

\usepackage{graphicx}
\usepackage{multirow}
\usepackage{arydshln}
\usepackage{enumitem}
\usepackage{ragged2e}

\title{SpecSteer: Synergizing Local Context and Global Reasoning for Efficient Personalized Generation }

\author{
 \textbf{Hang Lv\textsuperscript{1*}}
 \textbf{Sheng Liang\textsuperscript{2*}}, 
 \textbf{Hao Wang\textsuperscript{1}},
 \textbf{Yongyue Zhang\textsuperscript{2}},
 \textbf{Hongchao Gu\textsuperscript{1}},
 \\
 \textbf{Wei Guo\textsuperscript{2}},
 \textbf{Defu Lian\textsuperscript{1}},
 \textbf{Yong Liu\textsuperscript{2}},
 \textbf{Enhong Chen\textsuperscript{1}}
 \\
\textsuperscript{1}University of Science and Technology of China, \textsuperscript{2}Huawei Technologies Co., Ltd.
\\
}

\begin{document}
\maketitle


\begin{abstract}

Realizing personalized intelligence faces a core dilemma: sending user history to centralized large language models raises privacy concerns, while on-device small language models lack the reasoning capacity required for high-quality generation. Our pilot study shows that purely local enhancements remain insufficient to reliably bridge this gap. We therefore propose \textsc{SpecSteer}, an asymmetric collaborative inference framework that synergizes private on-device context with cloud-scale reasoning. 
\textsc{SpecSteer} casts collaboration as Bayesian knowledge fusion and repurposes speculative decoding as a distributed alignment protocol, yielding a \emph{Draft--Verify--Recover} pipeline: the on-device model drafts personalized sequences; the cloud validates via a ratio-based mechanism that decouples reasoning verification from private context, filtering logical flaws without accessing raw user context; upon rejection, a steering recovery injects local intent during correction.
Experiments demonstrate that \textsc{SpecSteer} successfully closes the reasoning gap and achieves superior personalized generation performance, while delivering a 2.36$\times$ speedup over standard baselines.

\end{abstract}


\section{Introduction}

The evolution of Large Language Models (LLMs) is shifting the focus from generic conversation to personalized assistance \cite{yu2025thoughtaugmentedplanningllmpoweredinteractive,zhang2026paradigmusercentricagentplatformcentric}. Realizing this potential with centralized architectures remains challenging, as effective personalization typically requires uploading sensitive user history to the cloud, raising concerns regarding data exposure and latency bottlenecks. To address these constraints, there is a growing trend toward deploying lightweight models directly on edge devices~\cite{gunter2024appleintelligencefoundationlanguage,seed2025seed15thinkingadvancingsuperbreasoning}, seeking to balance data sovereignty with the demand for rapid, personalized responses.

While this decentralized approach improves privacy and efficiency, it raises a fundamental question about performance: can these compact models truly synthesize complex user context into high-quality responses? To investigate this, we conducted a systematic pilot study (Section~\ref{sec:pilot}). The results reveal a persistent gap: even when equipped with advanced enhancement techniques, personalized small models fail to surpass generic large models—despite the latter having no access to the user's private context. 
This confirms a significant capacity deficit, where the informational advantage of local data is negated by limited \textbf{reasoning ability}, indicating that local optimization alone cannot achieve the sophisticated generation quality required for intelligent personalized applications.

\begin{figure}[t]
    \centering
    \includegraphics[width=0.5\textwidth]{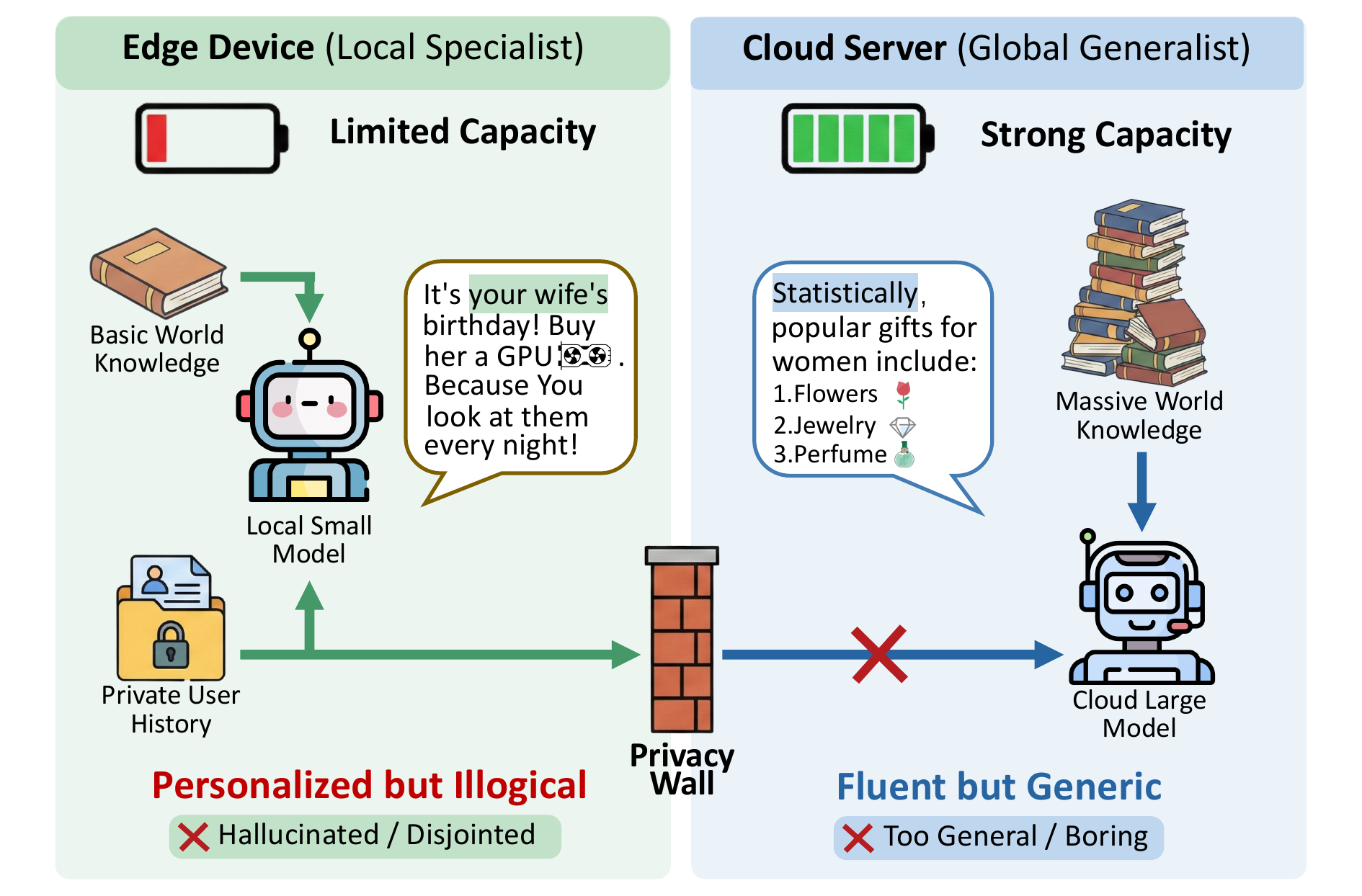}
    
    \caption{The fundamental dilemma in personalized generation: the inevitable trade-off between the reasoning capability required for high-quality responses and the data isolation necessary for privacy protection. }
    
    \label{fig:main_method}
\end{figure}
Ideally, we aim to construct a collaborative architecture that synergizes the superior reasoning capabilities of a large Generalist with the exclusive contextual access of a lightweight Specialist. However, realizing this vision encounters two fundamental hurdles.
The first is \textbf{information asymmetry}. Since the generalist lacks access to private data, bridging the two models is challenging. Sequential workflows that rely on discrete text exchange~\cite{frugal,LSRP} often result in \textit{coarse-grained alignment }, as the specialist's fine-grained intent is inevitably degraded when compressed into rigid prompts.
The second is \textbf{real-time latency}. Achieving tight collaboration requires frequent synchronization, yet the dense, step-by-step integration mandated by token-level fusion~\cite{proxytuning,cogenesis,costeer} introduces prohibitive communication overhead. This makes such methods impractical for interactive agents, creating a tension between the need for precise alignment and the constraint of inference speed.

To bridge this gap, we introduce \textsc{SpecSteer}, an efficient collaborative framework that combines local intent with global reasoning. We formulate collaborative inference as a Bayesian fusion process, where the cloud-based Generalist calibrates locally drafted candidates without accessing the private context behind them, and repurpose speculative decoding as a distributed alignment protocol to improve both personalized quality and inference latency. Specifically, the Local Specialist \textbf{drafts} personalized sequences using private history, and the Cloud Generalist performs \textbf{ratio-based verification}, validating logical plausibility against a generic (non-personalized) baseline without observing raw user data. When a draft is rejected, we perform \textbf{steering recovery} by biasing the recovery distribution with the Specialist's personalized signals, ensuring that corrected outputs remain faithful to user intent.
We validate the effectiveness of our framework through extensive experiments and discussions.
Our contributions are summarized as follows:
\begin{itemize}[leftmargin=1em,labelsep=0.4em,itemsep=0.2ex,topsep=0.3ex,parsep=0pt,partopsep=0pt]
    \item \textbf{Empirical Analysis:} We reveal the intrinsic capacity deficit of local models, demonstrating that purely local enhancement remains insufficient for robust personalized generation.
    \item \textbf{\textsc{SpecSteer} Framework:} We propose \textsc{SpecSteer}, an asymmetric edge-cloud framework that steers cloud-scale reasoning with local intent through a Bayesian \textit{Draft-Verify-Recover} protocol, preserving privacy while maintaining generation coherence.
    \item \textbf{Extensive Evaluation:} We demonstrate in extensive experiments that \textsc{SpecSteer} improves generation quality while achieving a 2.36$\times$ inference speedup, validating its practical effectiveness for edge-cloud personalized generation.
\end{itemize}

\section{Related Work}

\subsection{Personalized LLM}

Existing personalization methods mainly fall into two categories: Retrieval-Augmented Generation and Parameter-Efficient Fine-Tuning \cite{liu2025surveypersonalizedlargelanguage}. 
RAG-based methods retrieve user-specific context to guide generation \cite{huang-etal-2025-selfaug,gu-etal-2025-rapid}. MemoRAG \cite{memorag} uses a global memory module for long-term interactions, while RECAP \cite{recap} and ROPG \cite{ropg} improve context integration. 
PEFT-based methods instead adapt model weights to individual users. PocketLLM \cite{pocketllm} and OPPU \cite{oppu} study efficient on-device adaptation, while FDLORA \cite{fdlora} and PER-PCS \cite{perpcs} explore federated strategies for balancing quality and  efficiency. 

Despite their differences, these approaches share the same dilemma: \textbf{centralized methods} offer stronger reasoning but expose sensitive contexts or parameters, whereas \textbf{purely on-device methods} preserve privacy but are limited by lightweight model capacity. Our work addresses this tension through edge-cloud collaboration.

\subsection{Small-Large Model Collaboration}

Collaborative inference generally falls into two categories: sequential interaction and token-level fusion \cite{wang2025surveycollaboratingsmalllarge}. 
Sequential interaction relies on cascading pipelines where models communicate through discrete text exchanges. LSRP \cite{LSRP} aligns cloud guidance with local data via dynamic retrieval and small-model feedback, while CoWest \cite{cowest}, SuperICL \cite{supericl}, and CaLM \cite{calm} incorporate SLM predictions or verification signals to improve alignment. However, such coarse-grained interaction lacks step-wise supervision and can accumulate alignment drift in long-form generation. 
Token-level fusion, in contrast, combines model distributions at each decoding step. Proxy Tuning \cite{proxytuning} and CoGenesis \cite{cogenesis} steer LLMs with logit offsets or privacy-preserving sketches from SLMs, and CoSteer \cite{costeer} extends this idea to decoding-time personalization. However, these methods require \textit{synchronized inference}, introducing communication overhead that undermines real-time efficiency. 

Our work bridges this gap by providing fine-grained steering without requiring synchronization at every decoding step.

\begin{table*}[h]
\centering

\resizebox{\textwidth}{!}{
\begin{tabular}{clcccccccccccc}
\toprule
\multirow{2}{*}{\textbf{Model}} & \multirow{2}{*}{\textbf{Method}} & \multicolumn{2}{c}{\textbf{LaMP-4}} & \multicolumn{2}{c}{\textbf{LaMP-5}} & \multicolumn{2}{c}{\textbf{LaMP-7}} & \multicolumn{2}{c}{\textbf{Abstract}} & \multicolumn{2}{c}{\textbf{Review}} & \multicolumn{2}{c}{\textbf{Writing}} \\
\cmidrule(lr){3-4} \cmidrule(lr){5-6} \cmidrule(lr){7-8} \cmidrule(lr){9-10} \cmidrule(lr){11-12} \cmidrule(lr){13-14}
& & R1 & RL & R1 & RL & R1 & RL & R1 & RL & R1 & RL & R1 & RL \\
\midrule
\multirow{6}{*}{\rotatebox[origin=c]{90}{\textbf{Qwen3}}} 
& 0.6B Direct       & 12.29 & 10.60 & 44.98 & 37.13 & 45.89 & 40.43 & 36.58 & 20.34 & 24.15 & 12.95 & 26.32 & 12.72 \\
& LoRA              & 13.14 & 11.23 & 47.74 & 40.50 & 46.20 & 40.44 & 33.01 & 20.52 & 22.22 & 13.34 & 14.20 & 10.45 \\
& RAG               & 12.14 & 9.96  & 48.03 & 40.09 & 46.09 & 40.34 & 39.89 & 21.65 & 23.18 & 12.84 & 25.50 & 12.36 \\
& LoRA + RAG        & 10.36 & 9.26  & 44.88 & 38.71 & 46.16 & 40.36 & 35.04 & 21.43 & 24.42 & 14.69 & 16.91 & 10.47 \\
& RAFT              & 8.23  & 7.52  & 37.98 & 32.86 & 46.36 & 40.58 & 39.98 & 21.86 & 22.00 & 12.45 & 26.90 & \textbf{12.87} \\
\cdashline{2-14} 
& 32B Direct        & \textbf{14.48} & \textbf{12.34} & \textbf{48.71} & \textbf{40.98} & \textbf{48.06} & \textbf{42.00} & \textbf{40.18} & \textbf{22.17} & \textbf{31.18} & \textbf{14.78} & \textbf{29.46} & 12.64 \\
\midrule
\multirow{6}{*}{\rotatebox[origin=c]{90}{\textbf{Qwen2.5}}} 
& 1.5B Direct       & 12.13 & 10.12 & 44.94 & 38.49 & 32.68 & 28.88 & 36.97 & 17.60 & 22.50 & 10.86 & 23.83 & 10.38 \\
& LoRA              & 13.08 & 10.45 & 46.96 & \textbf{40.87} & 34.47 & 30.65 & 37.73 & 20.00 & 26.51 & 12.47 & 26.70 & 11.64 \\
& RAG               & 9.37  & 8.18  & 43.01 & 34.89 & 28.13 & 24.56 & 36.81 & 19.00 & 26.39 & 12.12 & 24.53 & 11.41 \\
& LoRA + RAG        & 12.11 & 10.20 & 47.72 & 38.71 & 29.82 & 26.41 & 34.64 & 19.29 & 28.76 & 12.55 & 20.93 & 10.53 \\
& RAFT              & 7.88  & 6.95  & 32.40 & 28.01 & 35.86 & 26.26 & 36.65 & 19.03 & 25.12 & 12.29 & 23.90 & 11.18 \\
\cdashline{2-14} 
& 32B Direct        & \textbf{13.64} & \textbf{11.23} & \textbf{48.58} & 39.71 & \textbf{38.52} & \textbf{33.35} & \textbf{38.50} & \textbf{20.51} & \textbf{31.73} & \textbf{14.41} & \textbf{29.14} & \textbf{12.20} \\
\bottomrule
\end{tabular}
}
\caption{\textbf{Empirical quantification of the Capacity Deficit.} We compare mobile-scale SLMs equipped with varying knowledge enhancement strategies (LoRA, RAG, RAFT) against standard server-scale Generalists (32B) across six personalized generative tasks. 
Metrics are Rouge-1\&L, the best results are marked in \textbf{bold}.}
\label{tab:pilot}
\end{table*}

\subsection{Speculative Decoding}

Speculative Decoding (SD) was originally introduced to accelerate inference through a "Draft-then-Verify" paradigm \cite{speculativedecoding,speculativesampling}. By using a lightweight drafter to approximate the target distribution, SD reduces wall-clock latency through selective verification. To improve acceptance rates, methods such as Medusa \cite{medusa} and Eagle \cite{eagle} introduce multi-head decoding or feature-level autoregression. Beyond acceleration, recent work has adapted the verification stage for controlled generation, aligning outputs with human preferences (SSS \cite{SSS}) or safety constraints (Judge Decoding \cite{bachmann2025judgedecodingfasterspeculative}). 

However, these methods predominantly operate under \textbf{Information Symmetry}, assuming the Verifier possesses full context. In our asymmetric setting, this assumption fails as the Cloud Verifier lacks access to private history, leading to erroneous rejections of personalized content. We fundamentally repurpose SD to bridge this gap, enabling the cloud to verify logical coherence without accessing the raw private data used by the local drafter.

\section{Pilot Study: The Capacity Deficit}
\label{sec:pilot}

In this section, we present an empirical study of the capability gap between local personalized agents and cloud-based generalists. Our goal is to test whether a mobile-scale model, even when enhanced with private knowledge via RAG or fine-tuning, can match a server-scale model without getting such access.

\subsection{Experiment Setup}

We conduct experiments on LaMP \cite{lamp} and LongLaMP \cite{longlamp}, two benchmarks for personalized generation.

\paragraph{Task Selection.} 
We focus on personalized generative tasks, including text summarization, email generation, and article writing, which require complex reasoning and the integration of user history into new content, making them more representative of realistic personalized agents. Specifically, we use six datasets across the two benchmarks.\footnote{We exclude LaMP-6 and LongLaMP-1 due to copyright restrictions on their underlying corpora.} Results on classification tasks (LaMP-1 to LaMP-3) are reported in Appendix \ref{sec:lamp13}.

\paragraph{Model Selection.}
To cover different scales, we use two representative model pairs, each contrasting a \textit{mobile-scale} specialist with a \textit{server-scale} generalist: Qwen3-0.6B vs. Qwen3-32B \cite{qwen3}, and Qwen2.5-1.5B-Instruct vs. Qwen2.5-32B-Instruct \cite{qwen2.5}.

\paragraph{Experimental Configurations.}
To test whether local enhancement can close the reasoning gap, we evaluate the SLMs under five progressively stronger settings: (A) \textit{Zero-shot}, (B) \textit{RAG}, (C) \textit{LoRA}, (D) \textit{LoRA + RAG}, and (E) \textit{RAFT} (Retrieval-Augmented Fine-Tuning + RAG). These fully enhanced local models are compared against server-scale LLMs in standard \textbf{Zero-shot} mode, allowing us to isolate the role of model capacity from information access. Detailed settings are provided in Appendix \ref{sec:setting}.

\newpage

\subsection{Results and Analysis}

The results in Table~\ref{tab:pilot} reveal a clear \textbf{reasoning ceiling} for small models. Despite lacking access to private user history, the server-scale Generalist outperforms fully enhanced local models on 22 of 24 metrics. This suggests that retrieval or fine-tuning cannot fully compensate for the limited capacity of small models to synthesize information into coherent reasoning.

Furthermore, local enhancement is also \textbf{highly inconsistent}: no single strategy, including RAG or PEFT, works reliably across tasks, indicating that fixed local optimization is inherently brittle.

Qualitatively, we observe a \textbf{fundamental tension} between contextual fidelity and logical robustness. Local Specialists better capture personalized intent and style but are more prone to contextual hallucination, whereas the Generalist remains structurally sound but defaults to a generic persona. 

Overall, these findings show that local optimization alone is insufficient, motivating \textsc{SpecSteer} to combine the Specialist's private grounding with the Generalist's reasoning strength.

\begin{figure*}[t]
    \centering
    \includegraphics[width=1.0\textwidth]{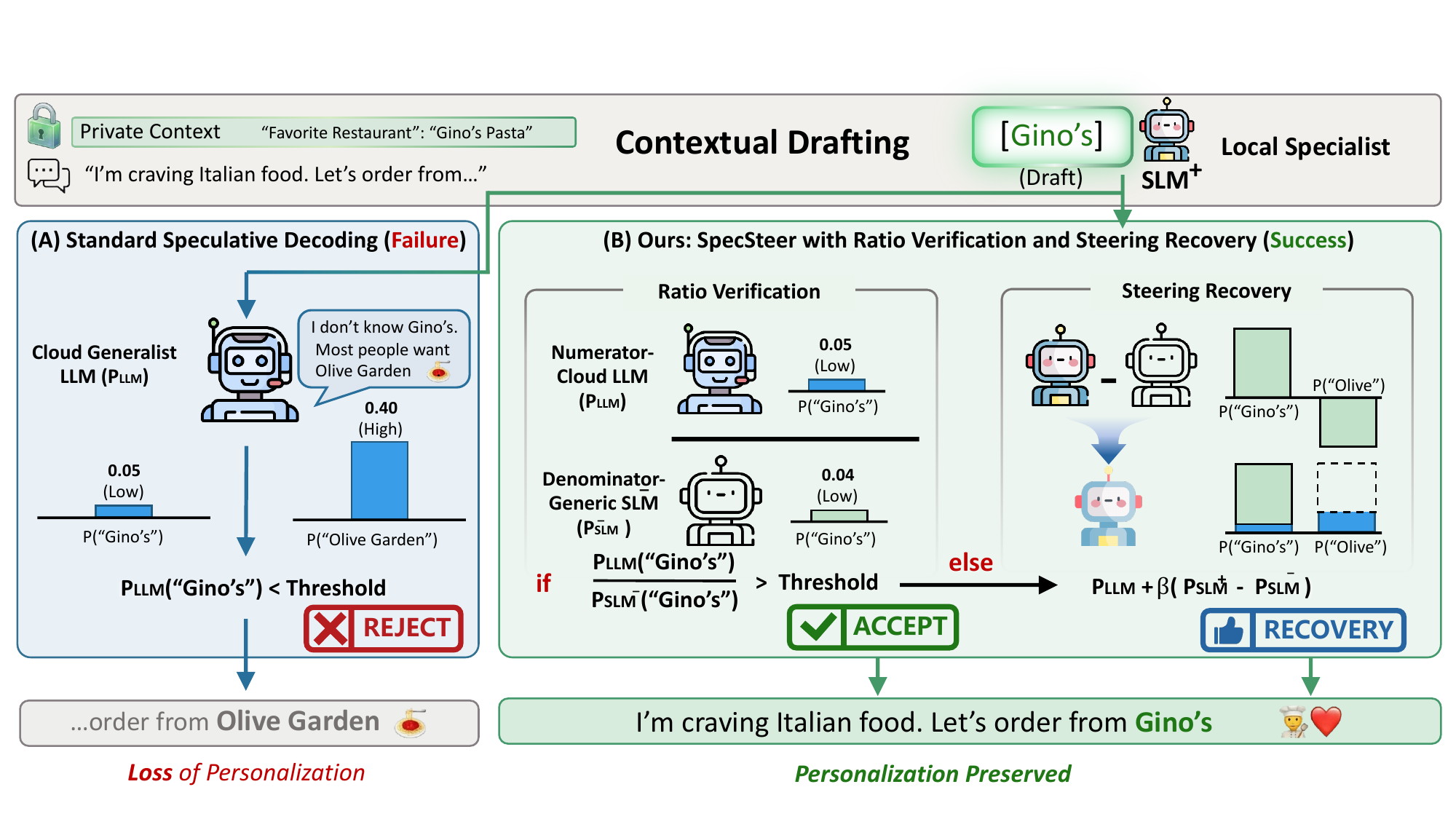}
    \caption{Intuitive illustration of \textbf{SpecSteer}. Standard speculative verification (A) may wrongly reject valid private entities (e.g., ``Gino's'') because the Generalist assigns them low absolute probability in the absence of private context. In contrast, SpecSteer (B) first performs \textbf{ratio-based verification}, normalizing the Generalist's score by a generic baseline ($P_{\mathrm{SLM}^{-}}$) to filter out surprise caused by missing private knowledge. Draft tokens that pass the ratio test are accepted directly, while failed cases are handled by \textbf{steering recovery}, which re-injects the Specialist's personalized signal to recover the intended continuation without exposing private data.}
    
    \label{fig:main_method}
\end{figure*}

\section{Methodology}

\label{sec:method}

We present \textsc{SpecSteer}, a collaborative inference framework designed for the asymmetric interaction between a large-scale \textit{Generalist} ($\mathcal{M}_{\text{LLM}}$) and a small-scale \textit{Specialist} ($\mathcal{M}_{\text{SLM}}$) possessing exclusive personalized knowledge. Rather than resorting to heuristic integration, we formulate this collaboration as a \textbf{Bayesian Knowledge Fusion} problem: approximating an ideal joint distribution that synergizes global reasoning capabilities with granular personalized intent. We demonstrate that this optimization objective can be efficiently solved by repurposing the speculative decoding paradigm, thereby achieving high-quality personalized generation with minimal latency. The overall workflow is illustrated in Alg~\ref{alg:specsteer}. 
We focus on the algorithmic derivation in this section, deferring physical deployment details to Appendix~\ref{app:deployment}.

\begin{algorithm}[t]
\caption{SpecSteer Framework}
\label{alg:specsteer}
\begin{algorithmic}[1]
\REQUIRE Generalist $\mathcal{M}_{\text{LLM}}$, Specialist $\mathcal{M}_{\text{SLM}}^\pm$, Context $y$, Horizon $K$, Params $\lambda, \beta$
\WHILE{not End-Of-Sequence}
    \STATE \textbf{// Phase 1: Contextual Drafting }
    \STATE Generate $K$ draft tokens $\hat{y}_{1:K}$ via $\mathcal{M}_{\text{SLM}}^+$ 
    \STATE \textbf{// Phase 2: Ratio-Based Verification }
    \STATE Compute logits $h_{\text{LLM}}$ and $h_{\text{SLM}}^-$ for sequence $[y, \hat{y}_{1:K}]$ in parallel
    
    \FOR{$t = 1$ \TO $K$}
        \STATE $\alpha_t \leftarrow \min\left(1, \frac{P_{\text{LLM}}(\hat{y}_t)}{\lambda \cdot P_{\text{SLM}}^-(\hat{y}_t)}\right)$ \COMMENT{Eq.~\ref{eq:verification_rule}}
        
        \IF{$\text{Uniform}(0,1) \le \alpha_t$}
            \STATE $y \leftarrow y \cup \hat{y}_t$ \COMMENT{Accept draft}
        \ELSE
            \STATE \textbf{// Phase 3: Steering Recovery}
            \STATE $h_{\text{rec}} \leftarrow h_{\text{LLM}}^{(t)} + \beta \cdot (h_{\text{SLM}}^{+(t)} - h_{\text{SLM}}^{-(t)})$ \COMMENT{Eq.~\ref{eq:contrastive_recovery}}
            \STATE Sample $y_{\text{new}} \sim \text{Softmax}(h_{\text{rec}})$
            \STATE $y \leftarrow y \cup y_{\text{new}}$
            \STATE \textbf{break}
        \ENDIF
    \ENDFOR

\ENDWHILE
\RETURN $y$
\end{algorithmic}
\end{algorithm}
\subsection{Problem Formulation}
\label{sec:problem_formulation}

\paragraph{Preliminaries.}

Let $x$ denote the user prompt. The generation proceeds autoregressively: at step $t$, the model predicts the next token $y_t \in \mathcal{V}$ given the history $y_{<t}$. We designate the Generalist as the \textit{prior distribution} $P_{\text{LLM}}(y_t | x, y_{<t})$. In our optimization framework, this acts as the reference policy ensuring adherence to global reasoning patterns. For brevity, we omit the explicit conditioning terms $(x, y_{<t})$ in the subsequent notation, denoting the probability simply as $P(y_t)$. Our objective is to derive an optimal policy $\pi^*(y_t)$ that refines this prior by incorporating personalized intent.


\paragraph{Bayesian Optimization Objective.}
To synthesize the collaborative distribution, we formulate the fusion as a step-wise constrained optimization problem. We seek a policy $\pi_t^*$ that maximizes the expected \textit{Personalized Intent} $r(y_t)$—derived from the specialist's private context—while minimizing the deviation from the generalist's prior to preserve reasoning consistency:
\begin{equation}
    \pi_t^* = \arg \max_{\pi_t} \left( \mathbb{E}_{y_t \sim \pi_t}[r(y_t)] - D_{\text{KL}}(\pi_t || P_{\text{LLM}}) \right)
    \label{eq:objective}
\end{equation}

\paragraph{Optimal Collaborative Distribution.}
By applying the method of Lagrange multipliers, we derive the closed-form solution to Eq.~\ref{eq:objective} as:
\begin{equation}
    \pi^*(y_t) = \frac{1}{Z_t} P_{\text{LLM}}(y_t) \exp(r(y_t))
    \label{eq:optimal_solution}
\end{equation}
where $Z_t$ acts as the partition function. The detailed derivation is provided in Appendix~\ref{app:derivation}.


\paragraph{Instantiating Personalized Intent via PMI.}
To operationalize the optimization, we must quantify the \textit{Personalized Intent} $r(y_t)$ defined in Eq.~\ref{eq:objective}. We interpret this term as the specific information gain attributable to the private user context. Following \citet{amulet} and \citet{costeer}, we posit that the personalized value of a token is measured not by its absolute likelihood, but by its divergence from a context-agnostic baseline.
Accordingly, we formulate this intent as the \textit{Pointwise Mutual Information (PMI)} provided by the private context:
\begin{equation}
    r(y_t) = \log P_{\text{SLM}}^+(y_t) - \log P_{\text{SLM}}^-(y_t)
    \label{eq:reward_definition}
\end{equation}
Here, $\mathcal{M}_{\text{SLM}}^+$ denotes the Specialist conditioned on private data, while $\mathcal{M}_{\text{SLM}}^-$ represents the generic baseline. 
Substituting Eq.~\ref{eq:reward_definition} into Eq.~\ref{eq:optimal_solution}, the exponential term $\exp(r(y_t))$ acts as a re-weighting factor, yielding the final target distribution $P^*$:
\begin{equation}
    P^*(y_t) \propto P_{\text{LLM}}(y_t) \cdot \frac{P_{\text{SLM}}^+(y_t)}{P_{\text{SLM}}^-(y_t)}
    \label{eq:final_target}
\end{equation}

\subsection{\textsc{SpecSteer} Framework}
\label{sec:speculative_steering}

Directly sampling from the collaborative target $P^*$ (Eq.~\ref{eq:final_target}) is infeasible because the constituent probability distributions are distributed across two distinct models. The Generalist lacks access to the private components ($P_{\text{SLM}}^+$), while the Specialist cannot compute the heavy global prior ($P_{\text{LLM}}$). 
To circumvent this deadlock, we recast the collaborative generation problem within the framework of speculative decoding~\cite{speculativedecoding,speculativesampling}. We repurpose the standard \textit{Draft-then-Verify} paradigm—traditionally serving as a latency optimization—transforming it into a distributed protocol for Bayesian alignment. This architectural shift decouples the generation process, allowing each model to contribute its specific expertise without requiring full state synchronization.

\subsubsection{Drafting: Contextual Proposal.}
We entrust the drafting role $Q(y_t)$ to the Specialist $\mathcal{M}_{\text{SLM}}^+$, capitalizing on its exclusive access to private history. Since the personalized intent is intrinsically local, the Specialist serves as the active driver of the generation trajectory. It efficiently constructs a draft hypothesis $\hat{y}$ that is pre-aligned with user constraints, effectively priming the global reasoning process:
\begin{equation}
    \hat{y} \sim Q(y_t) = P_{\text{SLM}}^+(y_t)
\end{equation}

\subsubsection{Verification: The Ratio-Based Protocol.}

The Generalist serves as the verifier, tasked with enforcing adherence to the collaborative target $P^*$. Unlike standard speculative decoding where the target is merely the LLM itself, here the acceptance criterion $\alpha = \min(1, P^*/Q)$ must account for the fused distribution.
Substituting the definition of $P^*$ (Eq.~\ref{eq:final_target}) and the drafter formulation $Q = P_{\text{SLM}}^+$ yields:
\begin{equation}
    \alpha(\hat{y}) = \min\left(1, \frac{P_{\text{LLM}}(\hat{y}) \cdot (P_{\text{SLM}}^+(\hat{y}) / P_{\text{SLM}}^-(\hat{y}))}{Z_t \cdot P_{\text{SLM}}^+(\hat{y})}\right)
\end{equation}
\textbf{This substitution reveals a critical structural advantage}: the private distribution $P_{\text{SLM}}^+$ appears in both the target numerator and the proposal denominator, cancelling out exactly. This mathematical property inherently decouples the verification process from the private context, ensuring the Generalist never requires access to raw sensitive data.

Finally, addressing the intractability of the partition function $Z_t$, we approximate it via a scalar hyperparameter $\lambda$. This leads to our computable \textbf{Ratio-Based Verification Rule}:
\begin{equation}
    \alpha(\hat{y}) = \min\left(1, \frac{P_{\text{LLM}}(\hat{y})}{\lambda \cdot P_{\text{SLM}}^-(\hat{y})}\right)
    \label{eq:verification_rule}
\end{equation}
Functionally, $\lambda$ acts as a reasoning gatekeeper: a larger $\lambda$ imposes a stricter logic constraint, prioritizing the Generalist's prior, while a smaller $\lambda$ relaxes the boundary, allowing more aggressive steering by the Specialist.

\subsubsection{Recovery: Steering via Logit Injection.}

When verification fails, the standard fallback strategy—sampling purely from the Generalist's prior $P_{\text{LLM}}$—is counterproductive. Since the Generalist is agnostic to private context, reverting to its raw distribution causes intent collapse, discarding the user's specific constraints.
To maintain the steering trajectory even during error correction, we introduce a recovery mechanism that approximates the collaborative target $P^*$ in the logit space.
By taking the logarithm of Eq.~\ref{eq:final_target}, the multiplicative fusion transforms into an additive intervention. Let $h(\cdot)$ denote the pre-softmax logits. We synthesize the recovery logits $h_{\text{rec}}$ as:
\begin{equation}
    h_{\text{rec}}(y_t) = h_{\text{LLM}}(y_t) + \beta \cdot \underbrace{\left( h_{\text{SLM}}^+(y_t) - h_{\text{SLM}}^-(y_t) \right)}_{\text{Contrastive Steering Vector}}
    \label{eq:contrastive_recovery}
\end{equation}
Upon rejection, we sample the next token $y_t \sim \text{Softmax}(h_{\text{rec}})$. 
Here, the term $(h_{\text{SLM}}^+ - h_{\text{SLM}}^-)$ isolates the pure direction of personalization. The hyperparameter $\beta$ acts as a contextual anchor: it ensures that even when the Generalist intervenes to fix logical flaws, the corrected output remains firmly tethered to the user's specific domain rather than drifting into genericism.

\section{Experiments}

\begin{table}
\centering
\footnotesize
\setlength{\tabcolsep}{3.15pt} 
\renewcommand{\arraystretch}{1.05} 
\begin{tabular}{c  l cc cc cc}
\toprule
\multirow{2}{*}{\textbf{Pair}} & \multirow{2}{*}{\textbf{Setting}} & \multicolumn{2}{c}{\textbf{Abs.}} & \multicolumn{2}{c}{\textbf{Rev.}} & \multicolumn{2}{c}{\textbf{Wri.}} \\
\cmidrule(lr){3-4} \cmidrule(lr){5-6} \cmidrule(lr){7-8}
& & R1 & RL & R1 & RL & R1 & RL \\
\midrule
\multirow{4}{*}{A} 
& SLM       & 36.58 & 20.34 & 24.15 & 12.95 & 26.32 & 12.72 \\
& SLM+      & 39.89 & 21.65 & 23.18 & 12.84 & 25.50 & 12.36 \\
& LLM       & 40.18 & 22.17 & 31.18 & 14.78 & 29.46 & 12.64 \\
& SpecSteer & \textbf{41.35} & \textbf{23.57} & \textbf{33.03} & \textbf{17.26} & \textbf{30.79} & \textbf{12.88} \\
\addlinespace[0.15ex]
\cmidrule(lr){1-8}
\addlinespace[0.15ex]
\multirow{4}{*}{B} 
& SLM       & 36.97 & 17.60 & 22.50 & 10.86 & 23.83 & 10.38 \\
& SLM+      & 36.81 & 19.00 & 26.39 & 12.12 & 24.53 & 11.41 \\
& LLM       & 38.50 & 20.51 & 31.73 & 14.41 & 29.14 & 12.20 \\
& SpecSteer & \textbf{39.73} & \textbf{21.06} & \textbf{32.70} & \textbf{15.44} & \textbf{30.22} & \textbf{12.82} \\
\addlinespace[0.15ex]
\cmidrule(lr){1-8}
\addlinespace[0.15ex]
\multirow{4}{*}{C} 
& SLM       & 39.85 & 21.31 & 30.43 & 14.54 & 28.99 & 12.94 \\
& SLM+      & 41.38 & 22.33 & 32.77 & 15.42 & 29.02 & 13.55 \\
& LLM       & 40.18 & 22.17 & 31.18 & 14.78 & \textbf{29.46} & 12.64 \\
& SpecSteer & \textbf{43.91} & \textbf{23.54} & \textbf{34.81} & \textbf{16.29} & 29.40 & \textbf{14.06} \\
\addlinespace[0.15ex]
\cmidrule(lr){1-8}
\addlinespace[0.15ex]
\multirow{4}{*}{D} 
& SLM       & 36.24 & 19.35 & 28.72 & 13.94 & 20.86 & 10.11 \\
& SLM+      & 37.89 & 20.36 & 30.53 & 14.52 & 23.14 & 11.23 \\
& LLM       & 39.04 & 21.23 & 31.71 & \textbf{15.08} & \textbf{27.84} & 12.41 \\
& SpecSteer & \textbf{39.54} & \textbf{21.38} & \textbf{31.88} & 14.68 & 24.29 & \textbf{12.44} \\
\bottomrule
\end{tabular}
\caption{Performance comparison on LongLaMP. SLM+ denotes the retrieval-augmented specialist. }
\label{tab:effectiveness_results}
\end{table} 

\label{sec:experiments}

In this section, we validate \textsc{SpecSteer} on the LongLaMP benchmark \cite{longlamp}, organizing the evaluation into two critical dimensions. First, we examine \textbf{Effectiveness}, verifying whether the framework can successfully synergize local intent with global reasoning to outperform both fully armed local specialists and standalone cloud generalists. Subsequently, we analyze \textbf{Efficiency}, demonstrating that this collaboration achieves significant latency reduction compared to standard baselines, making it viable for deployment.

\subsection{Effectiveness Evaluation}

\paragraph{Experimental Setup.}
We evaluate \textsc{SpecSteer} on LongLaMP \cite{longlamp}. Following the edge-cloud setting, the Specialist leverages private user data via retrieval augmentation, while the Generalist operates in a zero-shot manner. We consider four Specialist/Generalist pairings: (A) Qwen3 0.6B/32B, (B) Qwen2.5 1.5B/32B, (C) Qwen3 8B/32B \cite{qwen2.5, qwen3}, and (D) Llama-3.2 1B / Llama-3.1 8B \cite{grattafiori2024llama3herdmodels}, using the Instruct versions for the Qwen2.5 and Llama series. In Table \ref{tab:effectiveness_results}, we compare against two core baselines: \textbf{SLM+}, the retrieval-augmented Specialist, and \textbf{LLM}, the zero-shot Generalist. 

In Table \ref{tab:broader_baselines}, we further compare with retrieval-based methods (BM25 \cite{bm25}, bge-reranker \cite{chen2024bge}, PAG \cite{pag}), PEFT-based methods (TAM, OPPU \cite{oppu}), and alignment-based methods (PAD \cite{pad}, CoPE \cite{cope}) on the representative Pair A setting. These baselines are largely \emph{orthogonal} to \textsc{SpecSteer}, as they mainly improve the local Specialist and can therefore be viewed as alternative instantiations of $M^+_{SLM}$. As one concrete example, Appendix \ref{sec:lora} instantiates \textsc{SpecSteer} with a LoRA-based Specialist.

\begin{table}
\centering
\footnotesize
\setlength{\tabcolsep}{3.0pt}
\renewcommand{\arraystretch}{1.05}
\begin{tabular}{l l cc cc cc}
\toprule
\multirow{2}{*}{\textbf{Type}} & \multirow{2}{*}{\textbf{Method}} & \multicolumn{2}{c}{\textbf{Abs.}} & \multicolumn{2}{c}{\textbf{Rev.}} & \multicolumn{2}{c}{\textbf{Wri.}} \\
\cmidrule(lr){3-4} \cmidrule(lr){5-6} \cmidrule(lr){7-8}
& & R1 & RL & R1 & RL & R1 & RL \\
\midrule
\multirow{3}{*}{Ret.}
& BM25          & 39.89 & 21.65 & 23.18 & 12.84 & 25.50 & 12.36 \\
& BGE           & 40.13 & 21.76 & 25.35 & 13.08 & 23.48 & 12.06 \\
& PAG           & 38.17 & 20.14 & 25.58 & 13.65 & 28.20 & 12.10 \\
\addlinespace[0.15ex]

\addlinespace[0.15ex]
\cmidrule(lr){1-8}
\addlinespace[0.15ex]

\multirow{2}{*}{Peft}
& TAM           & 36.70 & 20.40 & 23.95 & 12.20 & 27.84 & 12.29 \\
& OPPU          & 38.23 & 21.27 & 24.52 & 12.24 & 27.53 & 12.26 \\
\addlinespace[0.15ex]

\addlinespace[0.15ex]
\cmidrule(lr){1-8}
\addlinespace[0.15ex]

\multirow{2}{*}{Align.}
& PAD           & 38.57 & 21.91 & 27.89 & 13.44 & 27.52 & 12.77 \\
& CoPE          & 39.44 & 22.56 & 28.54 & 14.28 & 28.17 & 12.54 \\
\addlinespace[0.15ex]

\addlinespace[0.15ex]
\cmidrule(lr){1-8}
\addlinespace[0.15ex]

Collab.
& Ours     & \textbf{41.35} & \textbf{23.57} & \textbf{33.03} & \textbf{17.26} & \textbf{30.79} & \textbf{12.88} \\
\bottomrule
\end{tabular}
\caption{Comparison with broader personalized generation baselines on LongLaMP under the representative Pair A setting. We report additional retrieval-based, PEFT-based, and alignment-based baselines beyond those in Table~2. \textbf{Bold} indicates the best results.}
\label{tab:broader_baselines}
\end{table}

\paragraph{Results and Analysis.}
As shown in Table \ref{tab:effectiveness_results}, \textsc{SpecSteer} consistently improves performance across all model scales and families. The gains are especially clear in the lightweight settings (Groups A and B), where the Specialist is more vulnerable to noisy retrieved context. For example, on the Qwen3 0.6B Review task, SLM+ achieves 23.18 R1, far below the zero-shot Generalist (31.18 R1), while \textsc{SpecSteer} raises the score to 33.03 R1 by filtering and correcting noisy local drafts. This benefit persists even with a stronger Specialist: in Group C, where the retrieval-augmented Qwen3 8B model is already competitive, \textsc{SpecSteer} still yields further gains, reaching 43.91 R1 on Abstract and 34.81 R1 on Review. Moreover, the gains on the Llama-based Pair D further confirm that \textsc{SpecSteer} is not tied to a single model family, but generalizes across architectures. These results show that \textsc{SpecSteer} not only rescues weak Specialists, but also improves strong ones through collaborative verification and recovery.

Table \ref{tab:broader_baselines} further shows that these gains are not limited to the two core baselines in Table \ref{tab:effectiveness_results}. As
shown in Table \ref{tab:broader_baselines}, \textsc{SpecSteer} outperforms retrieval-based, PEFT-based, and alignment-based baselines, with particularly large gains on the more reasoning-intensive Review task. This suggests that strengthening the local Specialist alone is insufficient to match the benefit of our collaborative framework. It also supports the claim that these methods are largely \emph{orthogonal} to \textsc{SpecSteer}: they improve the Specialist itself, whereas our gains come from the collaboration between a personalized drafter and a stronger verifier/recovery model. 

Overall, results in Table \ref{tab:effectiveness_results} and \ref{tab:broader_baselines} show that \textsc{SpecSteer} generalizes across diverse pairings and remains strong against a broader set of personalized generation baselines.

\begin{table*}[t]
\centering
\resizebox{\textwidth}{!}{
\begin{tabular}{l|ccc|ccc|ccc|cc}
\toprule
\multirow{2}{*}{\textbf{Method}} & \multicolumn{3}{c|}{\textbf{Abstract}} & \multicolumn{3}{c|}{\textbf{Review}} & \multicolumn{3}{c|}{\textbf{Writing}} & \multicolumn{2}{c}{\textbf{Efficiency}} \\
 & R-1 & R-L & $\alpha$(\%) & R-1 & R-L & $\alpha$(\%) & R-1 & R-L & $\alpha$(\%) & Speed & Speedup \\
\midrule
Vanilla LLM & 40.18 & 22.17 & - & 31.18 & 14.78 & - & 29.46 & 12.64 & - & 22.58 & 1.00$\times$ \\
CoSteer \cite{costeer} & \textbf{41.98} & \textbf{24.61} & - & 32.73 & 15.93 & - & \textbf{31.26} & \textbf{13.38} & - & 9.71 & 0.43$\times$ \\
LightCoSteer \cite{costeer} & 40.99 & 22.62 & - & 32.61 & 16.04 & - & 30.41 & 13.11 & - & 16.03 & 0.71$\times$ \\
Standard SD & 39.90 & 22.23 & 35.52 & 31.24 & 15.03 & 29.02 & 29.08 & 12.41 & 39.65 & 22.13 & 0.98$\times$ \\
\midrule
\textbf{SpecSteer} ($\lambda=0.5$) & 41.35 & 23.57 & 56.36 & \textbf{33.03} & \textbf{17.26} & 53.28 & 30.79 & 12.88 & 57.19 & 39.29 & 1.74$\times$ \\
\textbf{SpecSteer} ($\lambda=0.1$) & 41.16 & 22.75 & \textbf{73.79} & 32.38 & 16.80 & \textbf{81.46} & 29.85 & 12.33 & \textbf{86.16} & \textbf{53.29} & \textbf{2.36$\times$} \\
\bottomrule
\end{tabular}
}
\caption{Efficiency and quality trade-off analysis on the Qwen3-0.6B/32B pairing. \textit{Speed} denotes the generation throughput (tokens/s), and \textit{Speedup} is measured relative to the standalone LLM inference. $\alpha$ represents the average token acceptance rate. \textsc{SpecSteer} maintains high generation quality while achieving a $2.36\times$ acceleration, whereas existing steering methods often lead to a net slowdown due to their persistent per-token computational overhead.}
\label{tab:efficiency_results}
\end{table*}

\subsection{Efficiency Evaluation}

\paragraph{Experimental Setup.}
We next evaluate whether the effectiveness gains of \textsc{SpecSteer} come with favorable inference efficiency. Following the Pair A setting used above, we measure latency using the Qwen3-0.6B/32B pair. Since our goal here is to isolate the efficiency benefit of the interaction mechanism, we compare against three representative collaborative paradigms rather than the orthogonal local-enhancement baselines in Table~\ref{tab:broader_baselines}: (1) Iterative Fusion (CoSteer) \citep{costeer}, which performs expensive iterative optimization ($T=20$) at every decoding step; (2) Continuous Single-Step Fusion (LightCoSteer \cite{costeer}), a streamlined variant ($T=1$) that applies the steering vector (Eq.~\ref{eq:contrastive_recovery}) to every token; and (3) Standard Speculative Decoding (SD \cite{speculativedecoding,speculativesampling}), which lacks an explicit steering mechanism to bridge the distribution shift.

\paragraph{Efficiency Analysis.}
Table~\ref{tab:efficiency_results} shows that \textsc{SpecSteer} achieves the best trade-off between generation quality and latency. Compared with always-on steering methods, \textsc{SpecSteer} applies correction only when draft tokens are rejected, thereby avoiding the per-token fusion overhead of CoSteer and LightCoSteer. As a result, it matches the quality of continuous fusion while achieving a \textbf{2.36$\times$ speedup} under high acceptance rates. By contrast, standard speculative decoding reaches only $0.98\times$ speedup, showing that personalization-induced distribution mismatch substantially limits acceptance without explicit alignment. These results confirm that \textsc{SpecSteer} is not only more effective, but also a substantially more efficient mechanism.

\subsection{Robustness and Practicality Analysis}

\paragraph{Robustness to Noisy Specialists.} A natural concern is whether \textsc{SpecSteer} remains effective when the local Specialist produces weak or noisy drafts. To study this, we evaluate settings with manual noise injection, weak retrieval (BM25), and stronger retrieval (BGE) in Appendix~\ref{sec:noise_robustness}. Across all cases, \textsc{SpecSteer} consistently improves over the corresponding SLM+ baseline and remains competitive with, or superior to, the standalone cloud LLM. This suggests that the framework acts as a robust logical filter: even when local drafts are imperfect, the Generalist can still reject and repair problematic generations while preserving useful personalized signals. This robustness is central to our design, since practical edge specialists are often imperfect rather than fully reliable.

\paragraph{Cross-Architecture Deployment.}
We further examine whether \textsc{SpecSteer} remains applicable when the Specialist and Generalist come from different model families and use different tokenizers. As detailed in Appendix \ref{sec:cross_architecture}, the framework is architecture-agnostic: verification operates directly on the drafted text using each model's native tokenizer, while recovery is implemented in a shared byte-level space. Empirically, pairing a Qwen3-0.6B Specialist with a Llama-3.1-8B Generalist still yields consistent gains over both SLM+ and LLM, confirming that \textsc{SpecSteer} is not tied to a single model family and remains effective in realistic heterogeneous edge-cloud deployments.

\paragraph{Sensitivity of Hyperparameters.}
We analyze the steering strength $\beta$ and verification threshold $\lambda$ in Appendix \ref{sec:beta} and \ref{sec:lambda}, respectively. The results show that \textsc{SpecSteer} is stable across a broad range of $\beta$ values, maintaining consistent gains for $\beta \in [0.5, 2.0]$, while overly large values can degrade coherence by pushing generation too far from the Generalist's prior. For $\lambda$, we observe the expected quality-efficiency trade-off: strict verification improves faithfulness to the Generalist distribution but limits acceptance, whereas a moderate range of $\lambda \in [0.1, 0.5]$ substantially increases the acceptance rate and unlocks most of the acceleration without noticeably harming generation quality.

\paragraph{Computational Cost.}
Beyond wall-clock latency, Appendix \ref{sec:compute} reports the total floating-point operations (FLOPs) of different strategies. Compared with standard LLM-RAG, which becomes increasingly expensive as context length grows, \textsc{SpecSteer} offloads private-context processing to the lightweight Specialist and invokes the large model only for verification and selective recovery. As a result, it achieves a nearly $3.5\times$ reduction in FLOPs in long-context scenarios, showing that the framework is not only faster in practice but also fundamentally more resource-efficient.
\section{Conclusion}
\label{sec:conclusion}

In this paper, we introduce \textsc{SpecSteer}, an asymmetric collaborative inference framework that reconciles local data constraints with cloud-based reasoning via a \textit{Draft-then-Verify} paradigm. Empirical evaluations show that \textsc{SpecSteer} improves personalized generation quality over competitive baselines while achieving a 2.36$\times$ speedup through efficient on-demand steering. These results position \textsc{SpecSteer} as a scalable and practical framework for real-world edge-cloud personalization.



\newpage
\section*{Limitations}
\label{sec:limitations}

As shown in the appendix, \textsc{SpecSteer} remains effective under noisy Specialists. Its gains narrow only when the local model no longer provides a reliable personalization signal; even then, the cloud Generalist can still maintain overall coherence, while the added value of local steering is reduced. In terms of privacy, \textsc{SpecSteer} keeps user context on device throughout inference. Moreover, it provides a modular foundation for integrating advanced privacy-preserving primitives to protect output tokens, without modifying the core \textit{Draft--Verify--Recover} protocol.


\bibliography{custom}

@misc{costeer,
      title={CoSteer: Collaborative Decoding-Time Personalization via Local Delta Steering}, 
      author={Hang Lv and Sheng Liang and Hao Wang and Hongchao Gu and Yaxiong Wu and Wei Guo and Defu Lian and Yong Liu and Enhong Chen},
      year={2025},
      eprint={2507.04756},
      archivePrefix={arXiv},
      primaryClass={cs.CL},
      url={https://arxiv.org/abs/2507.04756}, 
}

@inproceedings{cogenesis,
    title = "{C}o{G}enesis: A Framework Collaborating Large and Small Language Models for Secure Context-Aware Instruction Following",
    author = "Zhang, Kaiyan  and
      Wang, Jianyu  and
      Hua, Ermo  and
      Qi, Biqing  and
      Ding, Ning  and
      Zhou, Bowen",
    editor = "Ku, Lun-Wei  and
      Martins, Andre  and
      Srikumar, Vivek",
    booktitle = "Proceedings of the 62nd Annual Meeting of the Association for Computational Linguistics (Volume 1: Long Papers)",
    month = aug,
    year = "2024",
    address = "Bangkok, Thailand",
    publisher = "Association for Computational Linguistics",
    url = "https://aclanthology.org/2024.acl-long.235/",
    doi = "10.18653/v1/2024.acl-long.235",
    pages = "4295--4312"
}

@misc{proxytuning,
      title={Tuning Language Models by Proxy}, 
      author={Alisa Liu and Xiaochuang Han and Yizhong Wang and Yulia Tsvetkov and Yejin Choi and Noah A. Smith},
      year={2024},
      eprint={2401.08565},
      archivePrefix={arXiv},
      primaryClass={cs.CL},
      url={https://arxiv.org/abs/2401.08565}, 
}

@misc{gunter2024appleintelligencefoundationlanguage,
      title={Apple Intelligence Foundation Language Models}, 
      author={Tom Gunter and Zirui Wang and Chong Wang and Ruoming Pang and Andy Narayanan and Aonan Zhang and Bowen Zhang and Chen Chen and Chung-Cheng Chiu and David Qiu and Deepak Gopinath and Dian Ang Yap and Dong Yin and Feng Nan and Floris Weers and Guoli Yin and Haoshuo Huang and Jianyu Wang and Jiarui Lu and John Peebles and Ke Ye and Mark Lee and Nan Du and Qibin Chen and Quentin Keunebroek and Sam Wiseman and Syd Evans and Tao Lei and Vivek Rathod and Xiang Kong and Xianzhi Du and Yanghao Li and Yongqiang Wang and Yuan Gao and Zaid Ahmed and Zhaoyang Xu and Zhiyun Lu and Al Rashid and Albin Madappally Jose and Alec Doane and Alfredo Bencomo and Allison Vanderby and Andrew Hansen and Ankur Jain and Anupama Mann Anupama and Areeba Kamal and Bugu Wu and Carolina Brum and Charlie Maalouf and Chinguun Erdenebileg and Chris Dulhanty and Dominik Moritz and Doug Kang and Eduardo Jimenez and Evan Ladd and Fangping Shi and Felix Bai and Frank Chu and Fred Hohman and Hadas Kotek and Hannah Gillis Coleman and Jane Li and Jeffrey Bigham and Jeffery Cao and Jeff Lai and Jessica Cheung and Jiulong Shan and Joe Zhou and John Li and Jun Qin and Karanjeet Singh and Karla Vega and Kelvin Zou and Laura Heckman and Lauren Gardiner and Margit Bowler and Maria Cordell and Meng Cao and Nicole Hay and Nilesh Shahdadpuri and Otto Godwin and Pranay Dighe and Pushyami Rachapudi and Ramsey Tantawi and Roman Frigg and Sam Davarnia and Sanskruti Shah and Saptarshi Guha and Sasha Sirovica and Shen Ma and Shuang Ma and Simon Wang and Sulgi Kim and Suma Jayaram and Vaishaal Shankar and Varsha Paidi and Vivek Kumar and Xin Wang and Xin Zheng and Walker Cheng and Yael Shrager and Yang Ye and Yasu Tanaka and Yihao Guo and Yunsong Meng and Zhao Tang Luo and Zhi Ouyang and Alp Aygar and Alvin Wan and Andrew Walkingshaw and Andy Narayanan and Antonie Lin and Arsalan Farooq and Brent Ramerth and Colorado Reed and Chris Bartels and Chris Chaney and David Riazati and Eric Liang Yang and Erin Feldman and Gabriel Hochstrasser and Guillaume Seguin and Irina Belousova and Joris Pelemans and Karen Yang and Keivan Alizadeh Vahid and Liangliang Cao and Mahyar Najibi and Marco Zuliani and Max Horton and Minsik Cho and Nikhil Bhendawade and Patrick Dong and Piotr Maj and Pulkit Agrawal and Qi Shan and Qichen Fu and Regan Poston and Sam Xu and Shuangning Liu and Sushma Rao and Tashweena Heeramun and Thomas Merth and Uday Rayala and Victor Cui and Vivek Rangarajan Sridhar and Wencong Zhang and Wenqi Zhang and Wentao Wu and Xingyu Zhou and Xinwen Liu and Yang Zhao and Yin Xia and Zhile Ren and Zhongzheng Ren},
      year={2024},
      eprint={2407.21075},
      archivePrefix={arXiv},
      primaryClass={cs.AI},
      url={https://arxiv.org/abs/2407.21075}, 
}

@misc{seed2025seed15thinkingadvancingsuperbreasoning,
      title={Seed1.5-Thinking: Advancing Superb Reasoning Models with Reinforcement Learning}, 
      author={ByteDance Seed and : and Jiaze Chen and Tiantian Fan and Xin Liu and Lingjun Liu and Zhiqi Lin and Mingxuan Wang and Chengyi Wang and Xiangpeng Wei and Wenyuan Xu and Yufeng Yuan and Yu Yue and Lin Yan and Qiying Yu and Xiaochen Zuo and Chi Zhang and Ruofei Zhu and Zhecheng An and Zhihao Bai and Yu Bao and Xingyan Bin and Jiangjie Chen and Feng Chen and Hongmin Chen and Riwei Chen and Liangqiang Chen and Zixin Chen and Jinsong Chen and Siyan Chen and Kaiyuan Chen and Zhi Chen and Jin Chen and Jiecao Chen and Jinxin Chi and Weinan Dai and Ning Dai and Jiahui Dai and Shihan Dou and Yantao Du and Zhengyin Du and Jianhui Duan and Chen Dun and Ting-Han Fan and Jiazhan Feng and Junda Feng and Ziyuan Feng and Yuwei Fu and Wenqi Fu and Hanjie Fu and Hao Ge and Hongyi Guo and Mingji Han and Li Han and Wenhao Hao and Xintong Hao and Qianyu He and Jerry He and Feng He and Wen Heng and Zehua Hong and Qi Hou and Liang Hu and Shengding Hu and Nan Hu and Kai Hua and Qi Huang and Ziyue Huang and Hongzhi Huang and Zihao Huang and Ting Huang and Wenhao Huang and Wei Jia and Bin Jia and Xiaoying Jia and Yuhua Jiang and Haobin Jiang and Ziheng Jiang and Kaihua Jiang and Chengquan Jiang and Jianpeng Jiao and Xiaoran Jin and Xing Jin and Xunhao Lai and Zheng Li and Xiang Li and Liyi Li and Hongkai Li and Zheng Li and Shengxian Wan and Ya Wang and Yunshui Li and Chenggang Li and Niuniu Li and Siyu Li and Xi Li and Xiao Li and Aoyan Li and Yuntao Li and Nianning Liang and Xinnian Liang and Haibin Lin and Weijian Lin and Ye Lin and Zhicheng Liu and Guanlin Liu and Guanlin Liu and Chenxiao Liu and Yan Liu and Gaohong Liu and Juncai Liu and Chundian Liu and Deyi Liu and Kaibo Liu and Siyao Liu and Qi Liu and Yongfei Liu and Kang Liu and Gan Liu and Boyi Liu and Rui Long and Weiqiang Lou and Chenwei Lou and Xiang Luo and Yao Luo and Caiping Lv and Heyang Lv and Bole Ma and Qianli Ma and Hongzhi Ma and Yiyuan Ma and Jin Ma and Wenchang Ma and Tingting Ma and Chen Mao and Qiyang Min and Zhe Nan and Guanghan Ning and Jinxiang Ou and Haojie Pan and Renming Pang and Yanghua Peng and Tao Peng and Lihua Qian and Lihua Qian and Mu Qiao and Meng Qu and Cheng Ren and Hongbin Ren and Yong Shan and Wei Shen and Ke Shen and Kai Shen and Guangming Sheng and Jinlong Shi and Wenlei Shi and Guang Shi and Shuai Shuai Cao and Yuxin Song and Zuquan Song and Jing Su and Yifan Sun and Tao Sun and Zewei Sun and Borui Wan and Zihan Wang and Xiaohui Wang and Xi Wang and Shuguang Wang and Jun Wang and Qinlong Wang and Chenyuan Wang and Shuai Wang and Zihan Wang and Changbao Wang and Jiaqiang Wang and Shihang Wang and Xuwu Wang and Zaiyuan Wang and Yuxuan Wang and Wenqi Wang and Taiqing Wang and Chengzhi Wei and Houmin Wei and Ziyun Wei and Shufa Wei and Zheng Wu and Yonghui Wu and Yangjun Wu and Bohong Wu and Shuang Wu and Jingqiao Wu and Ning Wu and Shuangzhi Wu and Jianmin Wu and Chenguang Xi and Fan Xia and Yuqiao Xian and Liang Xiang and Boren Xiang and Bowen Xiao and Zhen Xiao and Xia Xiao and Yongsheng Xiao and Chao Xin and Shulin Xin and Yuwen Xiong and Jingjing Xu and Ziwen Xu and Chenyin Xu and Jiayi Xu and Yifan Xu and Wei Xu and Yufei Xu and Shikun Xu and Shipeng Yan and Shen Yan and Qingping Yang and Xi Yang and Tianhao Yang and Yuehang Yang and Yuan Yang and Ximing Yang and Zeyu Yang and Guang Yang and Yifan Yang and Xuesong Yao and Bairen Yi and Fan Yin and Jianian Yin and Ziqiang Ying and Xiangyu Yu and Hongli Yu and Song Yu and Menghan Yu and Huan Yu and Siyu Yuan and Jun Yuan and Yutao Zeng and Tianyang Zhan and Zheng Zhang and Yun Zhang and Mofan Zhang and Wang Zhang and Ru Zhang and Zhi Zhang and Tianqi Zhang and Xinyi Zhang and Zhexi Zhang and Sijun Zhang and Wenqiang Zhang and Xiangxiang Zhang and Yongtao Zhang and Yuyu Zhang and Ge Zhang and He Zhang and Yue Zhang and Renjie Zheng and Ningxin Zheng and Zhuolin Zheng and Yaowei Zheng and Chen Zheng and Xiaoyun Zhi and Wanjun Zhong and Cheng Zhong and Zheng Zhong and Baoquan Zhong and Xun Zhou and Na Zhou and Huan Zhou and Hang Zhu and Defa Zhu and Wenjia Zhu and Lei Zuo},
      year={2025},
      eprint={2504.13914},
      archivePrefix={arXiv},
      primaryClass={cs.CL},
      url={https://arxiv.org/abs/2504.13914}, 
}

@misc{wang2025surveycollaboratingsmalllarge,
      title={A Survey on Collaborating Small and Large Language Models for Performance, Cost-effectiveness, Cloud-edge Privacy, and Trustworthiness}, 
      author={Fali Wang and Jihai Chen and Shuhua Yang and Ali Al-Lawati and Linli Tang and Hui Liu and Suhang Wang},
      year={2025},
      eprint={2510.13890},
      archivePrefix={arXiv},
      primaryClass={cs.CL},
      url={https://arxiv.org/abs/2510.13890}, 
}

@misc{speculativedecoding,
      title={Fast Inference from Transformers via Speculative Decoding}, 
      author={Yaniv Leviathan and Matan Kalman and Yossi Matias},
      year={2023},
      eprint={2211.17192},
      archivePrefix={arXiv},
      primaryClass={cs.LG},
      url={https://arxiv.org/abs/2211.17192}, 
}

@misc{speculativesampling,
      title={Accelerating Large Language Model Decoding with Speculative Sampling}, 
      author={Charlie Chen and Sebastian Borgeaud and Geoffrey Irving and Jean-Baptiste Lespiau and Laurent Sifre and John Jumper},
      year={2023},
      eprint={2302.01318},
      archivePrefix={arXiv},
      primaryClass={cs.CL},
      url={https://arxiv.org/abs/2302.01318}, 
}

@misc{SSS,
      title={Reward-Shifted Speculative Sampling Is An Efficient Test-Time Weak-to-Strong Aligner}, 
      author={Bolian Li and Yanran Wu and Xinyu Luo and Ruqi Zhang},
      year={2025},
      eprint={2508.15044},
      archivePrefix={arXiv},
      primaryClass={cs.CL},
      url={https://arxiv.org/abs/2508.15044}, 
}

@misc{pag,
      title={Integrating Summarization and Retrieval for Enhanced Personalization via Large Language Models}, 
      author={Chris Richardson and Yao Zhang and Kellen Gillespie and Sudipta Kar and Arshdeep Singh and Zeynab Raeesy and Omar Zia Khan and Abhinav Sethy},
      year={2023},
      eprint={2310.20081},
      archivePrefix={arXiv},
      primaryClass={cs.CL},
      url={https://arxiv.org/abs/2310.20081}, 
}

@misc{memorag,
      title={MemoRAG: Boosting Long Context Processing with Global Memory-Enhanced Retrieval Augmentation}, 
      author={Hongjin Qian and Zheng Liu and Peitian Zhang and Kelong Mao and Defu Lian and Zhicheng Dou and Tiejun Huang},
      year={2025},
      eprint={2409.05591},
      archivePrefix={arXiv},
      primaryClass={cs.CL},
      url={https://arxiv.org/abs/2409.05591}, 
}

@misc{lora,
      title={LoRA: Low-Rank Adaptation of Large Language Models}, 
      author={Edward J. Hu and Yelong Shen and Phillip Wallis and Zeyuan Allen-Zhu and Yuanzhi Li and Shean Wang and Lu Wang and Weizhu Chen},
      year={2021},
      eprint={2106.09685},
      archivePrefix={arXiv},
      primaryClass={cs.CL},
      url={https://arxiv.org/abs/2106.09685}, 
}

@misc{liu2025surveypersonalizedlargelanguage,
      title={A Survey of Personalized Large Language Models: Progress and Future Directions}, 
      author={Jiahong Liu and Zexuan Qiu and Zhongyang Li and Quanyu Dai and Wenhao Yu and Jieming Zhu and Minda Hu and Menglin Yang and Tat-Seng Chua and Irwin King},
      year={2025},
      eprint={2502.11528},
      archivePrefix={arXiv},
      primaryClass={cs.AI},
      url={https://arxiv.org/abs/2502.11528}, 
}

@inproceedings{recap,
    title = "{RECAP}: Retrieval-Enhanced Context-Aware Prefix Encoder for Personalized Dialogue Response Generation",
    author = "Liu, Shuai  and
      Cho, Hyundong  and
      Freedman, Marjorie  and
      Ma, Xuezhe  and
      May, Jonathan",
    editor = "Rogers, Anna  and
      Boyd-Graber, Jordan  and
      Okazaki, Naoaki",
    booktitle = "Proceedings of the 61st Annual Meeting of the Association for Computational Linguistics (Volume 1: Long Papers)",
    month = jul,
    year = "2023",
    address = "Toronto, Canada",
    publisher = "Association for Computational Linguistics",
    url = "https://aclanthology.org/2023.acl-long.468/",
    doi = "10.18653/v1/2023.acl-long.468",
    pages = "8404--8419"
}

@misc{ropg,
      title={Optimization Methods for Personalizing Large Language Models through Retrieval Augmentation}, 
      author={Alireza Salemi and Surya Kallumadi and Hamed Zamani},
      year={2024},
      eprint={2404.05970},
      archivePrefix={arXiv},
      primaryClass={cs.CL},
      url={https://arxiv.org/abs/2404.05970}, 
}

@inproceedings{pocketllm,
    title = "{P}ocket{LLM}: Enabling On-Device Fine-Tuning for Personalized {LLM}s",
    author = "Peng, Dan  and
      Fu, Zhihui  and
      Wang, Jun",
    editor = "Habernal, Ivan  and
      Ghanavati, Sepideh  and
      Ravichander, Abhilasha  and
      Jain, Vijayanta  and
      Thaine, Patricia  and
      Igamberdiev, Timour  and
      Mireshghallah, Niloofar  and
      Feyisetan, Oluwaseyi",
    booktitle = "Proceedings of the Fifth Workshop on Privacy in Natural Language Processing",
    month = aug,
    year = "2024",
    address = "Bangkok, Thailand",
    publisher = "Association for Computational Linguistics",
    url = "https://aclanthology.org/2024.privatenlp-1.10/",
    pages = "91--96"
}

@inproceedings{oppu,
    title = "Democratizing Large Language Models via Personalized Parameter-Efficient Fine-tuning",
    author = "Tan, Zhaoxuan  and
      Zeng, Qingkai  and
      Tian, Yijun  and
      Liu, Zheyuan  and
      Yin, Bing  and
      Jiang, Meng",
    editor = "Al-Onaizan, Yaser  and
      Bansal, Mohit  and
      Chen, Yun-Nung",
    booktitle = "Proceedings of the 2024 Conference on Empirical Methods in Natural Language Processing",
    month = nov,
    year = "2024",
    address = "Miami, Florida, USA",
    publisher = "Association for Computational Linguistics",
    url = "https://aclanthology.org/2024.emnlp-main.372/",
    doi = "10.18653/v1/2024.emnlp-main.372",
    pages = "6476--6491"
}

@misc{fdlora,
      title={FDLoRA: Personalized Federated Learning of Large Language Model via Dual LoRA Tuning}, 
      author={Jiaxing QI and Zhongzhi Luan and Shaohan Huang and Carol Fung and Hailong Yang and Depei Qian},
      year={2024},
      eprint={2406.07925},
      archivePrefix={arXiv},
      primaryClass={cs.DC},
      url={https://arxiv.org/abs/2406.07925}, 
}

@inproceedings{perpcs,
    title = "Personalized Pieces: Efficient Personalized Large Language Models through Collaborative Efforts",
    author = "Tan, Zhaoxuan  and
      Liu, Zheyuan  and
      Jiang, Meng",
    editor = "Al-Onaizan, Yaser  and
      Bansal, Mohit  and
      Chen, Yun-Nung",
    booktitle = "Proceedings of the 2024 Conference on Empirical Methods in Natural Language Processing",
    month = nov,
    year = "2024",
    address = "Miami, Florida, USA",
    publisher = "Association for Computational Linguistics",
    url = "https://aclanthology.org/2024.emnlp-main.371/",
    doi = "10.18653/v1/2024.emnlp-main.371",
    pages = "6459--6475"
}

@misc{LSRP,
      title={LSRP: A Leader-Subordinate Retrieval Framework for Privacy-Preserving Cloud-Device Collaboration}, 
      author={Yingyi Zhang and Pengyue Jia and Xianneng Li and Derong Xu and Maolin Wang and Yichao Wang and Zhaocheng Du and Huifeng Guo and Yong Liu and Ruiming Tang and Xiangyu Zhao},
      year={2025},
      eprint={2505.05031},
      archivePrefix={arXiv},
      primaryClass={cs.IR},
      url={https://arxiv.org/abs/2505.05031}, 
}

@inproceedings{cowest,
    title = "Synergistic Weak-Strong Collaboration by Aligning Preferences",
    author = "Jiao, Yizhu  and
      Zhang, Xuchao  and
      Wang, Zhaoyang  and
      Ma, Yubo  and
      Deng, Zhun  and
      Wang, Rujia  and
      Bansal, Chetan  and
      Rajmohan, Saravan  and
      Han, Jiawei  and
      Yao, Huaxiu",
    editor = "Che, Wanxiang  and
      Nabende, Joyce  and
      Shutova, Ekaterina  and
      Pilehvar, Mohammad Taher",
    booktitle = "Proceedings of the 63rd Annual Meeting of the Association for Computational Linguistics (Volume 1: Long Papers)",
    month = jul,
    year = "2025",
    address = "Vienna, Austria",
    publisher = "Association for Computational Linguistics",
    url = "https://aclanthology.org/2025.acl-long.995/",
    doi = "10.18653/v1/2025.acl-long.995",
    pages = "20355--20371",
    ISBN = "979-8-89176-251-0"
}

@inproceedings{supericl,
    title = "Small Models are Valuable Plug-ins for Large Language Models",
    author = "Xu, Canwen  and
      Xu, Yichong  and
      Wang, Shuohang  and
      Liu, Yang  and
      Zhu, Chenguang  and
      McAuley, Julian",
    editor = "Ku, Lun-Wei  and
      Martins, Andre  and
      Srikumar, Vivek",
    booktitle = "Findings of the Association for Computational Linguistics: ACL 2024",
    month = aug,
    year = "2024",
    address = "Bangkok, Thailand",
    publisher = "Association for Computational Linguistics",
    url = "https://aclanthology.org/2024.findings-acl.18/",
    doi = "10.18653/v1/2024.findings-acl.18",
    pages = "283--294"
}

@inproceedings{calm,
    title = "{C}a{LM}: Contrasting Large and Small Language Models to Verify Grounded Generation",
    author = "Hsu, I-Hung  and
      Wang, Zifeng  and
      Le, Long  and
      Miculicich, Lesly  and
      Peng, Nanyun  and
      Lee, Chen-Yu  and
      Pfister, Tomas",
    editor = "Ku, Lun-Wei  and
      Martins, Andre  and
      Srikumar, Vivek",
    booktitle = "Findings of the Association for Computational Linguistics: ACL 2024",
    month = aug,
    year = "2024",
    address = "Bangkok, Thailand",
    publisher = "Association for Computational Linguistics",
    url = "https://aclanthology.org/2024.findings-acl.759/",
    doi = "10.18653/v1/2024.findings-acl.759",
    pages = "12782--12803"
}

@misc{medusa,
      title={Medusa: Simple LLM Inference Acceleration Framework with Multiple Decoding Heads}, 
      author={Tianle Cai and Yuhong Li and Zhengyang Geng and Hongwu Peng and Jason D. Lee and Deming Chen and Tri Dao},
      year={2024},
      eprint={2401.10774},
      archivePrefix={arXiv},
      primaryClass={cs.LG},
      url={https://arxiv.org/abs/2401.10774}, 
}

@misc{eagle,
      title={EAGLE: Speculative Sampling Requires Rethinking Feature Uncertainty}, 
      author={Yuhui Li and Fangyun Wei and Chao Zhang and Hongyang Zhang},
      year={2025},
      eprint={2401.15077},
      archivePrefix={arXiv},
      primaryClass={cs.LG},
      url={https://arxiv.org/abs/2401.15077}, 
}

@misc{bachmann2025judgedecodingfasterspeculative,
      title={Judge Decoding: Faster Speculative Sampling Requires Going Beyond Model Alignment}, 
      author={Gregor Bachmann and Sotiris Anagnostidis and Albert Pumarola and Markos Georgopoulos and Artsiom Sanakoyeu and Yuming Du and Edgar Schönfeld and Ali Thabet and Jonas Kohler},
      year={2025},
      eprint={2501.19309},
      archivePrefix={arXiv},
      primaryClass={cs.LG},
      url={https://arxiv.org/abs/2501.19309}, 
}

@misc{lamp,
      title={La{MP}: When Large Language Models Meet Personalization}, 
      author={Alireza Salemi and Sheshera Mysore and Michael Bendersky and Hamed Zamani},
      year={2023},
      eprint={2304.11406},
      archivePrefix={arXiv},
      primaryClass={cs.CL}
}

@misc{longlamp,
      title={LongLaMP: A Benchmark for Personalized Long-form Text Generation}, 
      author={Ishita Kumar and Snigdha Viswanathan and Sushrita Yerra and Alireza Salemi and Ryan A. Rossi and Franck Dernoncourt and Hanieh Deilamsalehy and Xiang Chen and Ruiyi Zhang and Shubham Agarwal and Nedim Lipka and Chien Van Nguyen and Thien Huu Nguyen and Hamed Zamani},
      year={2024},
      eprint={2407.11016},
      archivePrefix={arXiv},
      primaryClass={cs.CL},
      url={https://arxiv.org/abs/2407.11016}, 
}

@misc{comparing,
      title={Comparing Retrieval-Augmentation and Parameter-Efficient Fine-Tuning for Privacy-Preserving Personalization of Large Language Models}, 
      author={Alireza Salemi and Hamed Zamani},
      year={2025},
      eprint={2409.09510},
      archivePrefix={arXiv},
      primaryClass={cs.CL},
      url={https://arxiv.org/abs/2409.09510}, 
}

@article{qwen3,
    title={Qwen3 Technical Report}, 
    author={An Yang and Anfeng Li and Baosong Yang and Beichen Zhang and Binyuan Hui and Bo Zheng and Bowen Yu and Chang Gao and Chengen Huang and Chenxu Lv and Chujie Zheng and Dayiheng Liu and Fan Zhou and Fei Huang and Feng Hu and Hao Ge and Haoran Wei and Huan Lin and Jialong Tang and Jian Yang and Jianhong Tu and Jianwei Zhang and Jianxin Yang and Jiaxi Yang and Jing Zhou and Jingren Zhou and Junyang Lin and Kai Dang and Keqin Bao and Kexin Yang and Le Yu and Lianghao Deng and Mei Li and Mingfeng Xue and Mingze Li and Pei Zhang and Peng Wang and Qin Zhu and Rui Men and Ruize Gao and Shixuan Liu and Shuang Luo and Tianhao Li and Tianyi Tang and Wenbiao Yin and Xingzhang Ren and Xinyu Wang and Xinyu Zhang and Xuancheng Ren and Yang Fan and Yang Su and Yichang Zhang and Yinger Zhang and Yu Wan and Yuqiong Liu and Zekun Wang and Zeyu Cui and Zhenru Zhang and Zhipeng Zhou and Zihan Qiu},
    journal = {arXiv preprint arXiv:2505.09388},
    year={2025}
}

@article{qwen2.5,
    title   = {Qwen2.5 Technical Report}, 
    author  = {An Yang and Baosong Yang and Beichen Zhang and Binyuan Hui and Bo Zheng and Bowen Yu and Chengyuan Li and Dayiheng Liu and Fei Huang and Haoran Wei and Huan Lin and Jian Yang and Jianhong Tu and Jianwei Zhang and Jianxin Yang and Jiaxi Yang and Jingren Zhou and Junyang Lin and Kai Dang and Keming Lu and Keqin Bao and Kexin Yang and Le Yu and Mei Li and Mingfeng Xue and Pei Zhang and Qin Zhu and Rui Men and Runji Lin and Tianhao Li and Tingyu Xia and Xingzhang Ren and Xuancheng Ren and Yang Fan and Yang Su and Yichang Zhang and Yu Wan and Yuqiong Liu and Zeyu Cui and Zhenru Zhang and Zihan Qiu},
    journal = {arXiv preprint arXiv:2412.15115},
    year    = {2024}
}

@misc{grattafiori2024llama3herdmodels,
      title={The Llama 3 Herd of Models}, 
      author={Aaron Grattafiori and Abhimanyu Dubey and Abhinav Jauhri and Abhinav Pandey and Abhishek Kadian and Ahmad Al-Dahle and Aiesha Letman and Akhil Mathur and Alan Schelten and Alex Vaughan and Amy Yang and Angela Fan and Anirudh Goyal and Anthony Hartshorn and Aobo Yang and Archi Mitra and Archie Sravankumar and Artem Korenev and Arthur Hinsvark and Arun Rao and Aston Zhang and Aurelien Rodriguez and Austen Gregerson and Ava Spataru and Baptiste Roziere and Bethany Biron and Binh Tang and Bobbie Chern and Charlotte Caucheteux and Chaya Nayak and Chloe Bi and Chris Marra and Chris McConnell and Christian Keller and Christophe Touret and Chunyang Wu and Corinne Wong and Cristian Canton Ferrer and Cyrus Nikolaidis and Damien Allonsius and Daniel Song and Danielle Pintz and Danny Livshits and Danny Wyatt and David Esiobu and Dhruv Choudhary and Dhruv Mahajan and Diego Garcia-Olano and Diego Perino and Dieuwke Hupkes and Egor Lakomkin and Ehab AlBadawy and Elina Lobanova and Emily Dinan and Eric Michael Smith and Filip Radenovic and Francisco Guzmán and Frank Zhang and Gabriel Synnaeve and Gabrielle Lee and Georgia Lewis Anderson and Govind Thattai and Graeme Nail and Gregoire Mialon and Guan Pang and Guillem Cucurell and Hailey Nguyen and Hannah Korevaar and Hu Xu and Hugo Touvron and Iliyan Zarov and Imanol Arrieta Ibarra and Isabel Kloumann and Ishan Misra and Ivan Evtimov and Jack Zhang and Jade Copet and Jaewon Lee and Jan Geffert and Jana Vranes and Jason Park and Jay Mahadeokar and Jeet Shah and Jelmer van der Linde and Jennifer Billock and Jenny Hong and Jenya Lee and Jeremy Fu and Jianfeng Chi and Jianyu Huang and Jiawen Liu and Jie Wang and Jiecao Yu and Joanna Bitton and Joe Spisak and Jongsoo Park and Joseph Rocca and Joshua Johnstun and Joshua Saxe and Junteng Jia and Kalyan Vasuden Alwala and Karthik Prasad and Kartikeya Upasani and Kate Plawiak and Ke Li and Kenneth Heafield and Kevin Stone and Khalid El-Arini and Krithika Iyer and Kshitiz Malik and Kuenley Chiu and Kunal Bhalla and Kushal Lakhotia and Lauren Rantala-Yeary and Laurens van der Maaten and Lawrence Chen and Liang Tan and Liz Jenkins and Louis Martin and Lovish Madaan and Lubo Malo and Lukas Blecher and Lukas Landzaat and Luke de Oliveira and Madeline Muzzi and Mahesh Pasupuleti and Mannat Singh and Manohar Paluri and Marcin Kardas and Maria Tsimpoukelli and Mathew Oldham and Mathieu Rita and Maya Pavlova and Melanie Kambadur and Mike Lewis and Min Si and Mitesh Kumar Singh and Mona Hassan and Naman Goyal and Narjes Torabi and Nikolay Bashlykov and Nikolay Bogoychev and Niladri Chatterji and Ning Zhang and Olivier Duchenne and Onur Çelebi and Patrick Alrassy and Pengchuan Zhang and Pengwei Li and Petar Vasic and Peter Weng and Prajjwal Bhargava and Pratik Dubal and Praveen Krishnan and Punit Singh Koura and Puxin Xu and Qing He and Qingxiao Dong and Ragavan Srinivasan and Raj Ganapathy and Ramon Calderer and Ricardo Silveira Cabral and Robert Stojnic and Roberta Raileanu and Rohan Maheswari and Rohit Girdhar and Rohit Patel and Romain Sauvestre and Ronnie Polidoro and Roshan Sumbaly and Ross Taylor and Ruan Silva and Rui Hou and Rui Wang and Saghar Hosseini and Sahana Chennabasappa and Sanjay Singh and Sean Bell and Seohyun Sonia Kim and Sergey Edunov and Shaoliang Nie and Sharan Narang and Sharath Raparthy and Sheng Shen and Shengye Wan and Shruti Bhosale and Shun Zhang and Simon Vandenhende and Soumya Batra and Spencer Whitman and Sten Sootla and Stephane Collot and Suchin Gururangan and Sydney Borodinsky and Tamar Herman and Tara Fowler and Tarek Sheasha and Thomas Georgiou and Thomas Scialom and Tobias Speckbacher and Todor Mihaylov and Tong Xiao and Ujjwal Karn and Vedanuj Goswami and Vibhor Gupta and Vignesh Ramanathan and Viktor Kerkez and Vincent Gonguet and Virginie Do and Vish Vogeti and Vítor Albiero and Vladan Petrovic and Weiwei Chu and Wenhan Xiong and Wenyin Fu and Whitney Meers and Xavier Martinet and Xiaodong Wang and Xiaofang Wang and Xiaoqing Ellen Tan and Xide Xia and Xinfeng Xie and Xuchao Jia and Xuewei Wang and Yaelle Goldschlag and Yashesh Gaur and Yasmine Babaei and Yi Wen and Yiwen Song and Yuchen Zhang and Yue Li and Yuning Mao and Zacharie Delpierre Coudert and Zheng Yan and Zhengxing Chen and Zoe Papakipos and Aaditya Singh and Aayushi Srivastava and Abha Jain and Adam Kelsey and Adam Shajnfeld and Adithya Gangidi and Adolfo Victoria and Ahuva Goldstand and Ajay Menon and Ajay Sharma and Alex Boesenberg and Alexei Baevski and Allie Feinstein and Amanda Kallet and Amit Sangani and Amos Teo and Anam Yunus and Andrei Lupu and Andres Alvarado and Andrew Caples and Andrew Gu and Andrew Ho and Andrew Poulton and Andrew Ryan and Ankit Ramchandani and Annie Dong and Annie Franco and Anuj Goyal and Aparajita Saraf and Arkabandhu Chowdhury and Ashley Gabriel and Ashwin Bharambe and Assaf Eisenman and Azadeh Yazdan and Beau James and Ben Maurer and Benjamin Leonhardi and Bernie Huang and Beth Loyd and Beto De Paola and Bhargavi Paranjape and Bing Liu and Bo Wu and Boyu Ni and Braden Hancock and Bram Wasti and Brandon Spence and Brani Stojkovic and Brian Gamido and Britt Montalvo and Carl Parker and Carly Burton and Catalina Mejia and Ce Liu and Changhan Wang and Changkyu Kim and Chao Zhou and Chester Hu and Ching-Hsiang Chu and Chris Cai and Chris Tindal and Christoph Feichtenhofer and Cynthia Gao and Damon Civin and Dana Beaty and Daniel Kreymer and Daniel Li and David Adkins and David Xu and Davide Testuggine and Delia David and Devi Parikh and Diana Liskovich and Didem Foss and Dingkang Wang and Duc Le and Dustin Holland and Edward Dowling and Eissa Jamil and Elaine Montgomery and Eleonora Presani and Emily Hahn and Emily Wood and Eric-Tuan Le and Erik Brinkman and Esteban Arcaute and Evan Dunbar and Evan Smothers and Fei Sun and Felix Kreuk and Feng Tian and Filippos Kokkinos and Firat Ozgenel and Francesco Caggioni and Frank Kanayet and Frank Seide and Gabriela Medina Florez and Gabriella Schwarz and Gada Badeer and Georgia Swee and Gil Halpern and Grant Herman and Grigory Sizov and Guangyi and Zhang and Guna Lakshminarayanan and Hakan Inan and Hamid Shojanazeri and Han Zou and Hannah Wang and Hanwen Zha and Haroun Habeeb and Harrison Rudolph and Helen Suk and Henry Aspegren and Hunter Goldman and Hongyuan Zhan and Ibrahim Damlaj and Igor Molybog and Igor Tufanov and Ilias Leontiadis and Irina-Elena Veliche and Itai Gat and Jake Weissman and James Geboski and James Kohli and Janice Lam and Japhet Asher and Jean-Baptiste Gaya and Jeff Marcus and Jeff Tang and Jennifer Chan and Jenny Zhen and Jeremy Reizenstein and Jeremy Teboul and Jessica Zhong and Jian Jin and Jingyi Yang and Joe Cummings and Jon Carvill and Jon Shepard and Jonathan McPhie and Jonathan Torres and Josh Ginsburg and Junjie Wang and Kai Wu and Kam Hou U and Karan Saxena and Kartikay Khandelwal and Katayoun Zand and Kathy Matosich and Kaushik Veeraraghavan and Kelly Michelena and Keqian Li and Kiran Jagadeesh and Kun Huang and Kunal Chawla and Kyle Huang and Lailin Chen and Lakshya Garg and Lavender A and Leandro Silva and Lee Bell and Lei Zhang and Liangpeng Guo and Licheng Yu and Liron Moshkovich and Luca Wehrstedt and Madian Khabsa and Manav Avalani and Manish Bhatt and Martynas Mankus and Matan Hasson and Matthew Lennie and Matthias Reso and Maxim Groshev and Maxim Naumov and Maya Lathi and Meghan Keneally and Miao Liu and Michael L. Seltzer and Michal Valko and Michelle Restrepo and Mihir Patel and Mik Vyatskov and Mikayel Samvelyan and Mike Clark and Mike Macey and Mike Wang and Miquel Jubert Hermoso and Mo Metanat and Mohammad Rastegari and Munish Bansal and Nandhini Santhanam and Natascha Parks and Natasha White and Navyata Bawa and Nayan Singhal and Nick Egebo and Nicolas Usunier and Nikhil Mehta and Nikolay Pavlovich Laptev and Ning Dong and Norman Cheng and Oleg Chernoguz and Olivia Hart and Omkar Salpekar and Ozlem Kalinli and Parkin Kent and Parth Parekh and Paul Saab and Pavan Balaji and Pedro Rittner and Philip Bontrager and Pierre Roux and Piotr Dollar and Polina Zvyagina and Prashant Ratanchandani and Pritish Yuvraj and Qian Liang and Rachad Alao and Rachel Rodriguez and Rafi Ayub and Raghotham Murthy and Raghu Nayani and Rahul Mitra and Rangaprabhu Parthasarathy and Raymond Li and Rebekkah Hogan and Robin Battey and Rocky Wang and Russ Howes and Ruty Rinott and Sachin Mehta and Sachin Siby and Sai Jayesh Bondu and Samyak Datta and Sara Chugh and Sara Hunt and Sargun Dhillon and Sasha Sidorov and Satadru Pan and Saurabh Mahajan and Saurabh Verma and Seiji Yamamoto and Sharadh Ramaswamy and Shaun Lindsay and Shaun Lindsay and Sheng Feng and Shenghao Lin and Shengxin Cindy Zha and Shishir Patil and Shiva Shankar and Shuqiang Zhang and Shuqiang Zhang and Sinong Wang and Sneha Agarwal and Soji Sajuyigbe and Soumith Chintala and Stephanie Max and Stephen Chen and Steve Kehoe and Steve Satterfield and Sudarshan Govindaprasad and Sumit Gupta and Summer Deng and Sungmin Cho and Sunny Virk and Suraj Subramanian and Sy Choudhury and Sydney Goldman and Tal Remez and Tamar Glaser and Tamara Best and Thilo Koehler and Thomas Robinson and Tianhe Li and Tianjun Zhang and Tim Matthews and Timothy Chou and Tzook Shaked and Varun Vontimitta and Victoria Ajayi and Victoria Montanez and Vijai Mohan and Vinay Satish Kumar and Vishal Mangla and Vlad Ionescu and Vlad Poenaru and Vlad Tiberiu Mihailescu and Vladimir Ivanov and Wei Li and Wenchen Wang and Wenwen Jiang and Wes Bouaziz and Will Constable and Xiaocheng Tang and Xiaojian Wu and Xiaolan Wang and Xilun Wu and Xinbo Gao and Yaniv Kleinman and Yanjun Chen and Ye Hu and Ye Jia and Ye Qi and Yenda Li and Yilin Zhang and Ying Zhang and Yossi Adi and Youngjin Nam and Yu and Wang and Yu Zhao and Yuchen Hao and Yundi Qian and Yunlu Li and Yuzi He and Zach Rait and Zachary DeVito and Zef Rosnbrick and Zhaoduo Wen and Zhenyu Yang and Zhiwei Zhao and Zhiyu Ma},
      year={2024},
      eprint={2407.21783},
      archivePrefix={arXiv},
      primaryClass={cs.AI},
      url={https://arxiv.org/abs/2407.21783}, 
}

@misc{opad,
      title={On-the-fly Preference Alignment via Principle-Guided Decoding}, 
      author={Mingye Zhu and Yi Liu and Lei Zhang and Junbo Guo and Zhendong Mao},
      year={2025},
      eprint={2502.14204},
      archivePrefix={arXiv},
      primaryClass={cs.CL},
      url={https://arxiv.org/abs/2502.14204}, 
}

@article{bm25,
url = {http://dx.doi.org/10.1561/1500000019},
year = {2009},
volume = {3},
journal = {Foundations and Trends® in Information Retrieval},
title = {The Probabilistic Relevance Framework: BM25 and Beyond},
doi = {10.1561/1500000019},
issn = {1554-0669},
number = {4},
pages = {333-389},
author = {Stephen Robertson and Hugo Zaragoza}
}

@misc{vonwerra2022trl,
  author = {Leandro von Werra and Younes Belkada and Lewis Tunstall and Edward Beeching and Tristan Thrush and Nathan Lambert and Shengyi Huang and Kashif Rasul and Quentin Gallouédec},
  title = {TRL: Transformer Reinforcement Learning},
  year = {2020},
  publisher = {GitHub},
  journal = {GitHub repository},
  howpublished = {\url{https://github.com/huggingface/trl}}
}

@misc{amulet,
      title={Amulet: ReAlignment During Test Time for Personalized Preference Adaptation of LLMs}, 
      author={Zhaowei Zhang and Fengshuo Bai and Qizhi Chen and Chengdong Ma and Mingzhi Wang and Haoran Sun and Zilong Zheng and Yaodong Yang},
      year={2025},
      eprint={2502.19148},
      archivePrefix={arXiv},
      primaryClass={cs.CL},
      url={https://arxiv.org/abs/2502.19148}, 
}

@misc{frugal,
      title={FrugalGPT: How to Use Large Language Models While Reducing Cost and Improving Performance}, 
      author={Lingjiao Chen and Matei Zaharia and James Zou},
      year={2023},
      eprint={2305.05176},
      archivePrefix={arXiv},
      primaryClass={cs.LG},
      url={https://arxiv.org/abs/2305.05176}, 
}

@misc{flop,
      title={Scaling Laws for Neural Language Models}, 
      author={Jared Kaplan and Sam McCandlish and Tom Henighan and Tom B. Brown and Benjamin Chess and Rewon Child and Scott Gray and Alec Radford and Jeffrey Wu and Dario Amodei},
      year={2020},
      eprint={2001.08361},
      archivePrefix={arXiv},
      primaryClass={cs.LG},
      url={https://arxiv.org/abs/2001.08361}, 
}

@misc{hayase2025samplinglanguagemodelbyte,
      title={Sampling from Your Language Model One Byte at a Time}, 
      author={Jonathan Hayase and Alisa Liu and Noah A. Smith and Sewoong Oh},
      year={2025},
      eprint={2506.14123},
      archivePrefix={arXiv},
      primaryClass={cs.CL},
      url={https://arxiv.org/abs/2506.14123}, 
}

@misc{chen2024bge,
      title={BGE M3-Embedding: Multi-Lingual, Multi-Functionality, Multi-Granularity Text Embeddings Through Self-Knowledge Distillation}, 
      author={Jianlv Chen and Shitao Xiao and Peitian Zhang and Kun Luo and Defu Lian and Zheng Liu},
      year={2024},
      eprint={2402.03216},
      archivePrefix={arXiv},
      primaryClass={cs.CL}
}

@misc{pad,
      title={PAD: Personalized Alignment of LLMs at Decoding-Time}, 
      author={Ruizhe Chen and Xiaotian Zhang and Meng Luo and Wenhao Chai and Zuozhu Liu},
      year={2025},
      eprint={2410.04070},
      archivePrefix={arXiv},
      primaryClass={cs.CL},
      url={https://arxiv.org/abs/2410.04070}, 
}

@misc{cope,
      title={Personalized LLM Decoding via Contrasting Personal Preference}, 
      author={Hyungjune Bu and Chanjoo Jung and Minjae Kang and Jaehyung Kim},
      year={2025},
      eprint={2506.12109},
      archivePrefix={arXiv},
      primaryClass={cs.CL},
      url={https://arxiv.org/abs/2506.12109}, 
}

@inproceedings{huang-etal-2025-selfaug,
    title = "{S}elf{A}ug: Mitigating Catastrophic Forgetting in Retrieval-Augmented Generation via Distribution Self-Alignment",
    author = "Huang, Yuqing  and
      Zhang, Rongyang  and
      Wang, Qimeng  and
      Lu, Chengqiang  and
      Gao, Yan  and
      Yiwu  and
      Hu, Yao  and
      Zhi, Xuyang  and
      Liu, Guiquan  and
      Li, Xin  and
      Wang, Hao  and
      Chen, Enhong",
    editor = "Christodoulopoulos, Christos  and
      Chakraborty, Tanmoy  and
      Rose, Carolyn  and
      Peng, Violet",
    booktitle = "Findings of the Association for Computational Linguistics: EMNLP 2025",
    month = nov,
    year = "2025",
    address = "Suzhou, China",
    publisher = "Association for Computational Linguistics",
    url = "https://aclanthology.org/2025.findings-emnlp.763/",
    doi = "10.18653/v1/2025.findings-emnlp.763",
    pages = "14175--14190",
    ISBN = "979-8-89176-335-7"
}

@inproceedings{gu-etal-2025-rapid,
    title = "{RAPID}: Efficient Retrieval-Augmented Long Text Generation with Writing Planning and Information Discovery",
    author = "Gu, Hongchao  and
      Li, Dexun  and
      Dong, Kuicai  and
      Zhang, Hao  and
      Lv, Hang  and
      Wang, Hao  and
      Lian, Defu  and
      Liu, Yong  and
      Chen, Enhong",
    editor = "Che, Wanxiang  and
      Nabende, Joyce  and
      Shutova, Ekaterina  and
      Pilehvar, Mohammad Taher",
    booktitle = "Findings of the Association for Computational Linguistics: ACL 2025",
    month = jul,
    year = "2025",
    address = "Vienna, Austria",
    publisher = "Association for Computational Linguistics",
    url = "https://aclanthology.org/2025.findings-acl.859/",
    doi = "10.18653/v1/2025.findings-acl.859",
    pages = "16742--16763",
    ISBN = "979-8-89176-256-5"
}

@misc{yu2025thoughtaugmentedplanningllmpoweredinteractive,
      title={Thought-Augmented Planning for LLM-Powered Interactive Recommender Agent}, 
      author={Haocheng Yu and Yaxiong Wu and Hao Wang and Wei Guo and Yong Liu and Yawen Li and Yuyang Ye and Junping Du and Enhong Chen},
      year={2025},
      eprint={2506.23485},
      archivePrefix={arXiv},
      primaryClass={cs.CL},
      url={https://arxiv.org/abs/2506.23485}, 
}

@misc{zhang2026paradigmusercentricagentplatformcentric,
      title={The Next Paradigm Is User-Centric Agent, Not Platform-Centric Service}, 
      author={Luankang Zhang and Hang Lv and Qiushi Pan and Kefen Wang and Yonghao Huang and Xinrui Miao and Yin Xu and Wei Guo and Yong Liu and Hao Wang and Enhong Chen},
      year={2026},
      eprint={2602.15682},
      archivePrefix={arXiv},
      primaryClass={cs.IR},
      url={https://arxiv.org/abs/2602.15682}, 
}

\appendix

\section{Additional Results and Analysis}

\label{sec:addition_results}

\subsection{Results on LaMP 1-3}
\label{sec:lamp13}
On these simple discriminative tasks, local enhancements like LoRA or RAG are quite effective. However, their performance falls far short of large cloud models on generative tasks as shown in Table~\ref{tab:pilot}, highlighting a gap that simple local optimization cannot bridge.

\begin{table}[htbp]
\centering
\setlength{\tabcolsep}{3pt} 
\scriptsize 
\begin{tabular*}{\columnwidth}{@{\extracolsep{\fill}}lccccc}
\toprule
\multirow{2}{*}{\textbf{Method}} & \textbf{LaMP-1} & \multicolumn{2}{c}{\textbf{LaMP-2}} & \multicolumn{2}{c}{\textbf{LaMP-3}} \\
\cmidrule(lr){2-2} \cmidrule(lr){3-4} \cmidrule(lr){5-6}
& Acc $\uparrow$ & Acc $\uparrow$ & F1 $\uparrow$ & MAE $\downarrow$ & RMSE $\downarrow$ \\
\midrule
\multicolumn{6}{l}{\textbf{Qwen3-0.6B}} \\
\midrule
Direct Gen.       & 0.46 & 0.02 & 0.0119 & 1.67 & 2.1900 \\
LoRA              & 0.45 & 0.85 & 0.1577 & 0.43 & 0.8775 \\
RAG               & 0.48 & 0.62 & 0.1396 & 2.94 & 3.2100 \\
LoRA + RAG        & 0.53 & 0.70 & 0.1121 & 0.49 & 0.9767 \\
RAFT              & 0.59 & 0.49 & 0.1220 & 0.90 & 1.5700 \\
32B Direct Gen.   & 0.42 & 0.16 & 0.1003 & 0.41 & 0.6708 \\
\midrule
\multicolumn{6}{l}{\textbf{Qwen2.5-1.5B}} \\
\midrule
Direct Gen.       & 0.43 & 0.04 & 0.0377 & 0.42 & 0.7211 \\
LoRA              & 0.46 & 0.81 & 0.2532 & 0.29 & 0.5745 \\
RAG               & 0.50 & 0.03 & 0.0183 & 0.45 & 0.7681 \\
LoRA + RAG        & 0.53 & 0.73 & 0.2294 & 0.23 & 0.5000 \\
RAFT              & 0.52 & 0.49 & 0.2585 & 0.40 & 0.8000 \\
32B Results       & 0.36 & 0.16 & 0.1198 & 0.41 & 0.6708 \\
\bottomrule
\end{tabular*}
\caption{Performance Comparison of on LaMP 1-3. Indicators for LaMP-1 is Acc, LaMP-2 are Acc and F1, and LaMP-3 are MAE and RMSE.}
\label{tab:lamp13}
\end{table}

\subsection{SpecSteer with Lora}
\label{sec:lora}
In this section, we evaluate \textsc{SpecSteer} using LoRA as the local enhancement strategy, represented by Group E (Qwen3-0.6B/32B) and Group F (Qwen2.5-1.5B-Instruct/32B). The results in Table~\ref{tab:speclora} show that LoRA's impact on small models is quite inconsistent.
For Group E, the Qwen3-0.6B model's performance actually declines after LoRA training. Despite this, \textsc{SpecSteer} still produces better results than the SLM+ baseline, proving that the Generalist can effectively correct the logic of a \textit{degraded specialist}. In Group F, where LoRA successfully improves the Qwen2.5-1.5B model,  \textsc{SpecSteer} further boosts these gains to achieve the best overall performance. This confirms that our framework is robust and benefits from local LoRA tuning whenever it is effective.

\begin{table}[htbp]
\centering
\footnotesize
\setlength{\tabcolsep}{3.15pt} 
\renewcommand{\arraystretch}{1.05} 

\begin{tabular}{c l cc cc cc}
\toprule
\multirow{2}{*}{\textbf{Pair}} & \multirow{2}{*}{\textbf{Setting}} & \multicolumn{2}{c}{\textbf{Abs.}} & \multicolumn{2}{c}{\textbf{Rev.}} & \multicolumn{2}{c}{\textbf{Wri.}} \\
\cmidrule(lr){3-4} \cmidrule(lr){5-6} \cmidrule(lr){7-8}
& & R1 & RL & R1 & RL & R1 & RL \\
\midrule
\multirow{4}{*}{E} 
& SLM       & 36.58 & 20.34 & 24.15 & 12.95 & 26.32 & 12.72 \\
& SLM+      & 33.01 & 20.52 & 22.22 & 13.34 & 14.20 & 10.45 \\
& LLM       & \textbf{40.18} & \textbf{22.17} & \textbf{31.18} & \textbf{14.78} & \textbf{29.46} & \textbf{12.64} \\
& SpecSteer & 34.25 & 21.09 & 26.28 & 14.05 & 16.75 & 12.27 \\
\addlinespace[0.15ex]
\cmidrule(lr){1-8}
\addlinespace[0.15ex]
\multirow{4}{*}{F} 
& SLM       & 36.97 & 17.60 & 22.50 & 10.86 & 23.83 & 10.38 \\
& SLM+      & 37.73 & 20.00 & 26.51 & 12.47 & 26.70 & 11.64 \\
& LLM       & 38.50 & 20.51 & 31.73 & 14.41 & \textbf{29.14} & 12.20 \\
& SpecSteer & \textbf{39.46} & \textbf{21.34} & \textbf{32.87} & \textbf{14.85} & 28.13 & \textbf{12.91} \\
\bottomrule
\end{tabular}
\caption{Performance Comparison across Different Settings (Best results in \textbf{bold})}
\label{tab:speclora}
\end{table}

\subsection{Robustness to Noisy Specialists}
\label{sec:noise_robustness}

\begin{table}[htbp]
\centering
\footnotesize
\setlength{\tabcolsep}{2.5pt}
\renewcommand{\arraystretch}{1.1}
\begin{tabular}{l cc cc cc}
\toprule
\multirow{2}{*}{\textbf{Setting}} & \multicolumn{2}{c}{\textbf{Abstract}} & \multicolumn{2}{c}{\textbf{Review}} & \multicolumn{2}{c}{\textbf{Writing}} \\
\cmidrule(lr){2-3} \cmidrule(lr){4-5} \cmidrule(lr){6-7}
& R1 & RL & R1 & RL & R1 & RL \\
\midrule
SLM            & 36.58 & 20.34 & 24.15 & 12.95 & 26.32 & 12.72 \\
LLM            & 40.18 & 22.17 & 31.18 & 14.78 & 29.46 & 12.64 \\
\midrule
SLM+ (Noise)   & 37.07 & 21.12 & 23.54 & 12.58 & 24.99 & 12.03 \\
SLM+ (BM25)    & 39.89 & 21.65 & 23.18 & 12.84 & 25.50 & 12.36 \\
SLM+ (BGE)     & 40.13 & 21.76 & 25.35 & 13.08 & 23.48 & 12.06 \\
\midrule
SpecSteer (Noise) & 40.35 & 22.63 & 31.78 & 15.14 & 29.24 & \textbf{12.98} \\
SpecSteer (BM25)  & 41.35 & 23.57 & 33.03 & \textbf{17.26} & 30.79 & 12.88 \\
SpecSteer (BGE)   & \textbf{41.56} & \textbf{23.77} & \textbf{33.45} & 17.18 & \textbf{31.03} & 12.84 \\
\bottomrule
\end{tabular}
\caption{Robustness of \textsc{SpecSteer} under different levels of Specialist noise on Qwen3-0.6B/32B.}
\label{tab:noise_robustness}
\end{table}

A natural question is how strongly \textsc{SpecSteer} depends on the quality of the local Specialist. As discussed in Limitations, our framework cannot fully recover from extreme failure cases where the Specialist's foundational capability collapses after poor fine-tuning. However, such catastrophic degradation is different from the more common practical scenario in which the local model remains functional but produces weak or noisy drafts due to imperfect retrieval. To evaluate robustness under these more realistic conditions, we consider three settings with different levels of input quality: \textbf{(1) Noise}, where we manually append low-similarity context to the prompt; \textbf{(2) BM25}, which provides relatively weak retrieval; and \textbf{(3) BGE}, which provides stronger retrieval through a neural reranker.

Table~\ref{tab:noise_robustness} reports the corresponding results. Across all three settings, \textsc{SpecSteer} consistently outperforms the local Specialist and remains competitive with, or superior to, the standalone cloud LLM. The gains are especially clear on the Review task, where noisy local drafts are most harmful and the Generalist's verification plays the largest corrective role. For example, under manual noise injection, \textsc{SpecSteer} improves over SLM+ from 23.54 to 31.78 R1 on Review, effectively recovering most of the degradation caused by irrelevant context. Similar trends hold under both BM25 and BGE retrieval, indicating that the framework is robust not only to weak retrieval but also to imperfect draft quality more broadly. These results suggest that \textsc{SpecSteer} acts as a robust logical filter: even when the Specialist is suboptimal, the Generalist can reject and repair problematic drafts while still preserving useful personalized signals.

\subsection{Cross-Architecture Deployment}
\label{sec:cross_architecture}

\begin{table}
\centering
\footnotesize
\setlength{\tabcolsep}{5pt}
\renewcommand{\arraystretch}{1.1}
\begin{tabular}{l cc cc cc}
\toprule
\multirow{2}{*}{\textbf{Method}} & \multicolumn{2}{c}{\textbf{Abstract}} & \multicolumn{2}{c}{\textbf{Review}} & \multicolumn{2}{c}{\textbf{Writing}} \\
\cmidrule(lr){2-3} \cmidrule(lr){4-5} \cmidrule(lr){6-7}
& R1 & RL & R1 & RL & R1 & RL \\
\midrule
SLM        & 36.58 & 20.34 & 24.15 & 12.95 & 26.32 & 12.72 \\
SLM+       & 39.89 & 21.65 & 23.18 & 12.84 & 25.50 & 12.36 \\
LLM        & 39.04 & 21.23 & 31.71 & 15.08 & 27.84 & 12.41 \\
SpecSteer  & \textbf{41.23} & \textbf{23.75} & \textbf{32.03} & \textbf{15.24} & \textbf{29.57} & \textbf{13.61} \\
\bottomrule
\end{tabular}
\caption{Cross-architecture deployment results using Qwen3-0.6B as the Specialist and Llama-3.1-8B as the Generalist.}
\label{tab:cross_architecture}
\end{table}

A practical concern is whether \textsc{SpecSteer} remains applicable when the Specialist and Generalist come from different model families and thus use different tokenizers. Our framework is designed to be architecture-agnostic. During the \emph{Verify} stage, the two models score the drafted text segment using their own tokenizers, so no tokenizer alignment is required. During the \emph{Recovery} stage, we adopt a lightweight byte-level solution \cite{hayase2025samplinglanguagemodelbyte}: both models are mapped into a shared byte space before applying steering, which avoids any assumption of vocabulary overlap while preserving cross-model compatibility.

To validate this property, we pair a Qwen3-0.6B Specialist with a Llama-3.1-8B Generalist. Table~\ref{tab:cross_architecture} shows that \textsc{SpecSteer} still yields consistent gains over both SLM+ and LLM in this heterogeneous setting, improving performance across all three tasks. In particular, it raises Review from 23.18/12.84 (SLM+) and 31.71/15.08 (LLM) to 32.03/15.24, while also achieving the best Writing performance. These results show that \textsc{SpecSteer} is not tied to a single model family, and remains effective in realistic cross-architecture edge-cloud deployments.

\subsection{Sensitivity Analysis of Steering Strength}
\label{sec:beta}
We investigate the impact of the steering recovery parameter $\beta$ on the Qwen3-0.6B/32B pair. As shown in Table~\ref{tab:beta_study}, \textsc{SpecSteer} exhibits robust performance across a wide range of values ($\beta \in [0.5, 2.0]$), consistently outperforming the standalone LLM and the fully-armed SLM+ baseline.  However, once $\beta$ exceeds 2.5, we observe a noticeable decline in performance across all metrics. This is because excessive steering forces the generated tokens to deviate too far from the Generalist's original distribution, leading to disjointed sentences and logical incoherence. This suggests that while a strong personalization signal is beneficial, it must be balanced against the Generalist's logical guardrails to maintain high-quality generation.

\begin{table}
\centering
\footnotesize
\setlength{\tabcolsep}{5pt}
\renewcommand{\arraystretch}{1.1}
\begin{tabular}{l cc cc cc}
\toprule
\multirow{2}{*}{\textbf{Setting}} & \multicolumn{2}{c}{\textbf{Abstract}} & \multicolumn{2}{c}{\textbf{Review}} & \multicolumn{2}{c}{\textbf{Writing}} \\
\cmidrule(lr){2-3} \cmidrule(lr){4-5} \cmidrule(lr){6-7}
& R1 & RL & R1 & RL & R1 & RL \\
\midrule
SLM & 36.58 & 20.34 & 24.15 & 12.95 & 26.32 & 12.72 \\
SLM+ & 39.89 & 21.65 & 23.18 & 12.84 & 25.50 & 12.36 \\
LLM & 40.18 & 22.17 & 31.18 & 14.78 & 29.46 & 12.64 \\
\midrule
$\beta = 0.5$ & 41.55 & 22.67 & \textbf{33.43} & 16.76 & 30.89 & 12.18 \\
$\beta = 1.0$ & 41.35 & 23.57 & 33.03 & 17.26 & 30.79 & 12.88 \\
$\beta = 1.5$ & 40.15 & 24.67 & 32.13 & \textbf{17.36} & 29.49 & \textbf{13.48} \\
$\beta = 2.0$ & \textbf{41.95} & 22.47 & 32.83 & 16.96 & \textbf{31.99} & 12.08 \\
$\beta = 2.5$ & 38.85 & 21.25 & 30.50 & 14.85 & 28.20 & 11.45 \\
\bottomrule
\end{tabular}
\caption{Impact of Steering Strength $\beta$ on Qwen3-0.6B/32B.}
\label{tab:beta_study}
\end{table}

\subsection{Sensitivity Analysis of Verification Threshold}
\label{sec:lambda}

We further investigate the impact of the verification threshold $\lambda$ on the balance between generation quality and inference efficiency. As demonstrated in Table~\ref{tab:lambda_study}, $\lambda$ serves as a critical trade-off parameter: when set to a strict value of 1.0, the framework enforces rigorous adherence to the Generalist's distribution, yielding the highest ROUGE scores. However, this strictness causes the average acceptance rate to drop significantly—often below 40\%—which incurs frequent rejections and heavy communication overhead, resulting in a marginal speedup of only 1.12. Conversely, lowering $\lambda$ to 0.1 allows the acceptance rate to surge above 80\%, delivering a substantial 2.36x speedup. While efficiency continues to improve slightly as $\lambda$ drops to 0.01, the marginal gain in speed (only an additional 0.07x) is offset by a continuous decline in generation quality metrics. These results indicate that $\lambda \in [0.1, 0.5]$ represents the optimal operating window for \textsc{SpecSteer}. In this range, the framework effectively amortizes the computational cost of personalization by maintaining high acceptance rates while preserving the Generalist’s essential logical guardrails.

\begin{table*}
\centering
\footnotesize
\begin{tabular}{l ccc ccc ccc cc}
\toprule
\multirow{2}{*}{\textbf{Setting}} & \multicolumn{3}{c}{\textbf{Abstract}} & \multicolumn{3}{c}{\textbf{Review}} & \multicolumn{3}{c}{\textbf{Writing}} & \multicolumn{2}{c}{\textbf{Efficiency}} \\
\cmidrule(lr){2-4} \cmidrule(lr){5-7} \cmidrule(lr){8-10} \cmidrule(lr){11-12}
& R1 & RL & $\alpha$(\%) & R1 & RL & $\alpha$(\%) & R1 & RL & $\alpha$(\%) & Speed & Speedup \\
\midrule
$\lambda=1.0$ & 41.85 & 24.08 & 37.45 & 33.68 & 17.82 & 38.36 & 30.45 & 13.42 & 43.15 & 25.29 & 1.12x \\
$\lambda=0.5$ & 41.35 & 23.57 & 56.36 & 33.03 & 17.26 & 53.28 & 30.79 & 12.88 & 57.19 & 39.29 & 1.74x \\
$\lambda=0.1$ & 41.16 & 22.75 & 73.79 & 32.38 & 16.80 & 81.46 & 29.85 & 12.33 & 86.16 & 53.29 & 2.36x \\
$\lambda=0.01$ & 40.85 & 21.95 & 77.12 & 32.00 & 16.20 & 82.31 & 29.17 & 11.86 & 87.40 & 54.87 & 2.43x \\
\bottomrule
\end{tabular}
\caption{Impact of Verification Threshold $\lambda$ on Performance and Efficiency (Qwen3-0.6B/32B)}
\label{tab:lambda_study}
\end{table*}

\subsection{Computation Cost Analysis} 
\label{sec:compute}

To quantify the hardware efficiency of our framework, we analyze the total floating-point operations (FLOPs) required to generate 100 new tokens under varying context lengths \cite{flop}.  As illustrated in Table~\ref{tab:flops_analysis}, the traditional LLM-RAG approach suffers from a significant computational burden as the context size increases, with FLOPs surging from $4.01 \times 10^{12}$ to $6.64 \times 10^{12}$ when the context reaches 10,000 tokens. This escalation is primarily due to the quadratic complexity of the self-attention mechanism in large-scale models when processing long sequences. In contrast, \textsc{SpecSteer} significantly reduces the total computational cost by decoupling context processing from global reasoning. Since the Generalist (32B) in our framework does not require access to the raw, often lengthy private context, its computational overhead remains relatively flat even as the user's history grows. By offloading the context-heavy drafting task to the lightweight Specialist (0.6B) and leveraging speculative verification, \textsc{SpecSteer} ($\lambda=0.75$) achieves a nearly 3.5x reduction in FLOPs compared to the LLM-RAG baseline in long-context scenarios. These results demonstrate that \textsc{SpecSteer} is not only faster in terms of wall-clock time but also remarkably more efficient in resource utilization for real-world personalized agents.

\begin{table}[htbp]
\centering
\footnotesize
\setlength{\tabcolsep}{8pt}
\renewcommand{\arraystretch}{1.1}
\begin{tabular}{l ccc}
\toprule
\multirow{2}{*}{\textbf{Method}} & \multicolumn{3}{c}{\textbf{Context Length ($c$)}} \\
\cmidrule(lr){2-4}
& 100 & 1,000 & 10,000 \\
\midrule
LLM (RAG) & 4.01 & 4.25 & 6.64 \\
SpecSteer ($\lambda=0.5$) & 2.30 & 2.32 & 2.50 \\
SpecSteer ($\lambda=0.75$) & \textbf{1.72} & \textbf{1.74} & \textbf{1.88} \\
\bottomrule
\end{tabular}
\caption{Computational Cost Comparison (Total FLOPs $\times 10^{12}$) for Generating 100 Tokens}
\label{tab:flops_analysis}
\end{table}
\section{Tasks and Settings}

\label{sec:setting}
\subsection{Tasks}

We utilize LaMP \cite{lamp} and LongLaMP \cite{longlamp}, two prominent benchmarks in personalized generation, for our evaluation. The specific task categories, corresponding metrics, and illustrative samples are listed in Table \ref{tab:lamp_longlamp_final}.

\subsection{General Experiment Setting}
For all experiments, we use greedy decoding to ensure deterministic and reproducible results. The maximum generation length is set to 1,024 tokens. For the \textsc{SpecSteer} framework, we use a verification threshold $\lambda=0.5$ and a steering recovery strength $\beta=1.0$ as the default configuration across all tasks, unless otherwise specified in the sensitivity analysis.

\subsection{Pilot Experiment Setting}
Following previous work \cite{comparing}, we utilized the temporal split version for all tasks to strictly adhere to the chronological order of user interactions. To focus on users with sufficient data for personalization, we selected the top-100 most active users (based on interaction frequency) for our experiments.

To simulate a real-world scenario where Small Language Models (SLMs) are updated periodically rather than in real-time, we adopted a specific data partitioning strategy: for the training phase, we excluded the most recent 10 interactions for each user and utilized the remaining historical data for fine-tuning. However, for the retrieval phase, we utilized the entire user profile as the candidate corpus, ensuring the retrieval system has access to the most comprehensive context. The specific implementation details for RAG and LoRA are described in the following sections.

\subsection{RAG Setting}
Following previous work \cite{lamp,comparing,oppu}, we employ BM25 \cite{bm25} as the retrieval module to augment the Large Language Model (LLM) with personalized context. Specifically, for each input query, we retrieve the top-$k$ (with $k=4$) most relevant documents from the user's historical profile. These retrieved entries are concatenated to construct the user profile, which is then integrated into the LLM's context window. The detailed organization of prompts and template formats for each task are presented in Table \ref{tab:prompts}.

\subsection{Peft Setting}
Following previous work\cite{comparing,oppu}, we organized the training data for the personalization tasks as illustrated in Figure \ref{tab:peft_implementation}. We employed Low-Rank Adaptation (LoRA) \cite{lora} to fine-tune the base model. Specifically, we injected trainable adapters into all linear layers (including query, key, value, output, gate, up, and down projections) with a rank of $r=64$, a scaling factor $\alpha=32$, and a dropout rate of $0.05$. The models were trained for 5 epochs using a learning rate of $1\times 10^{-4}$ with a cosine decay scheduler. To ensure training efficiency, we set the maximum sequence length to 1024 tokens and utilized BF16 precision throughout the training process \cite{vonwerra2022trl}.

\section{Derivation of the Optimal Token-Level Policy}
\label{app:derivation}

In this section, we provide the detailed derivation of the optimal solution for the constrained optimization problem formulated in Eq.~\ref{eq:objective}.

\subsection{Problem Setup}

We seek to find the optimal policy distribution $\pi^*(y_t | x, y_{<t})$ at time step $t$ that maximizes the expected steering reward while maintaining minimal KL-divergence from the Generalist's prior $P_{\text{LLM}}(y_t | x, y_{<t})$. For brevity, let us simplify the notation as:
\begin{itemize}
    \item $\pi(y) \equiv \pi(y_t | x, y_{<t})$: The target policy to be optimized.
    \item $P(y) \equiv P_{\text{LLM}}(y_t | x, y_{<t})$: The fixed prior distribution from the Generalist.
    \item $r(y) \equiv r(y_t, x, y_{<t})$: The local steering reward.
\end{itemize}

The objective function $\mathcal{J}(\pi)$ is defined as:
\begin{equation}
    \mathcal{J}(\pi) = \mathbb{E}_{y \sim \pi}[r(y)] - D_{\text{KL}}(\pi || P)
\end{equation}
Expanding the terms, we have:
\begin{equation}
    \mathcal{J}(\pi) = \sum_{y \in \mathcal{V}} \pi(y) r(y) - \sum_{y \in \mathcal{V}} \pi(y) \log \frac{\pi(y)}{P(y)}
\end{equation}
We must optimize this objective subject to the probability constraint that the distribution sums to 1:
\begin{equation}
    \sum_{y \in \mathcal{V}} \pi(y) = 1
\end{equation}

\subsection{Lagrange Multiplier Method}

We introduce a Lagrange multiplier $\lambda$ to enforce the normalization constraint. The Lagrangian $\mathcal{L}$ is given by:
\begin{equation}
\begin{aligned}
    \mathcal{L}(\pi, \lambda) = & \sum_{y} \pi(y) r(y) - \sum_{y} \pi(y) \log \frac{\pi(y)}{P(y)} \\
    & + \lambda \left( 1 - \sum_{y} \pi(y) \right)
\end{aligned}
\end{equation}

To find the optimal distribution, we take the functional derivative of $\mathcal{L}$ with respect to $\pi(y)$ and set it to zero. Note that $\frac{\partial}{\partial \pi} (\pi \log \pi) = 1 + \log \pi$.
\begin{equation}
    \frac{\partial \mathcal{L}}{\partial \pi(y)} = r(y) - \left( \log \frac{\pi(y)}{P(y)} + 1 \right) - \lambda = 0
\end{equation}

Rearranging the terms to isolate $\log \pi(y)$:
\begin{equation}
    \log \frac{\pi(y)}{P(y)} = r(y) - 1 - \lambda
\end{equation}
\begin{equation}
    \log \pi(y) = \log P(y) + r(y) - (1 + \lambda)
\end{equation}

Exponentiating both sides yields:
\begin{equation}
    \pi(y) = P(y) \exp(r(y)) \exp(-1 - \lambda)
\end{equation}

\subsection{Normalization and Final Form}

Since $\sum_{y} \pi(y) = 1$, we can solve for the constant term $\exp(-1 - \lambda)$. Let us define the partition function $Z$ as:
\begin{equation}
    Z = \sum_{y' \in \mathcal{V}} P(y') \exp(r(y'))
\end{equation}
Integrating the sum constraint:
\begin{equation}
    \sum_{y} P(y) \exp(r(y)) \exp(-1 - \lambda) = 1
\end{equation}
\begin{equation}
    \exp(-1 - \lambda) \cdot Z = 1 \implies \exp(-1 - \lambda) = \frac{1}{Z}
\end{equation}

Substituting this back into the expression for $\pi(y)$, we obtain the final closed-form solution corresponding to Eq.~\ref{eq:optimal_solution} in the main text:
\begin{equation}
    \pi^*(y_t | x, y_{<t}) = \frac{1}{Z_t} P_{\text{LLM}}(y_t | x, y_{<t}) \exp(r(y_t, x, y_{<t}))
\end{equation}
This derivation follows the standard principles of functional variational optimization under constraints~\cite{opad}.

\section{System Deployment and Implementation Details}
\label{app:deployment}

In this section, we detail the physical realization of \textsc{SpecSteer}, focusing on the hybrid deployment strategy that minimizes communication overhead while preserving user context privacy.

\subsection{Hybrid Model Placement}
To support the asymmetric collaboration described in Section~\ref{sec:method}, we distribute the model components across the edge and cloud as follows:

\begin{itemize}
    \item \textbf{Local Edge (User Side):} Hosts the \textbf{Personalized Specialist ($\mathcal{M}_{\text{SLM}}^+$)}. This model is fine-tuned on or has access to the user's private context $\mathcal{C}$ (e.g., retrieval indices, interaction history). It is responsible for the \textit{Contextual Drafting} phase and the final \textit{Steering Recovery} if needed.
    
    \item \textbf{Cloud Server (Service Side):} Hosts the \textbf{Generalist ($\mathcal{M}_{\text{LLM}}$)} and a generic, non-private copy of the \textbf{Specialist ($\mathcal{M}_{\text{SLM}}^-$)}. The $\mathcal{M}_{\text{SLM}}^-$ shares the same architecture and base weights as the local specialist but lacks access to user-specific contexts.
\end{itemize}

\subsection{Implementation of Cloud-Centric Verification}
A key challenge in distributed speculative decoding is the communication bottleneck: transmitting high-dimensional logit matrices from the drafter to the verifier is often prohibitively expensive. We address this by mathematically decoupling the verification step from local distributions.

Recall our target distribution derived in Eq.~(2): $P^* \propto P_{\text{LLM}} \cdot (P_{\text{SLM}}^+ / P_{\text{SLM}}^-)$. Standard speculative decoding accepts a token $\hat{y}$ from the drafter ($P_{\text{SLM}}^+$) with probability $\alpha = \min(1, \frac{P^*(\hat{y})}{P_{\text{SLM}}^+(\hat{y})})$.

Substituting $P^*$ into the acceptance criterion, the term $P_{\text{SLM}}^+(\hat{y})$ cancels out:
\begin{align}
\alpha &= \min\left(1, \frac{P_{\text{LLM}}(\hat{y}) \cdot \frac{P_{\text{SLM}}^+(\hat{y})}{P_{\text{SLM}}^-(\hat{y})}}{P_{\text{SLM}}^+(\hat{y})}\right)\\
&= \min\left(1, \frac{P_{\text{LLM}}(\hat{y})}{P_{\text{SLM}}^-(\hat{y})}\right)
\end{align}

This derivation leads to a highly efficient implementation workflow:
\begin{enumerate}
    \item \textbf{Token-Only Uplink:} The Local Edge generates a draft sequence $\hat{y}_{1:K}$ using $P_{\text{SLM}}^+$ and sends \textbf{only the token IDs} to the Cloud. Crucially, the edge \textit{does not} need to transmit its large probability distributions ($P_{\text{SLM}}^+$) or the private context.
    
    \item \textbf{Parallel Computation on Cloud:} Upon receiving $\hat{y}_{1:K}$, the Cloud runs $\mathcal{M}_{\text{LLM}}$ and $\mathcal{M}_{\text{SLM}}^-$ in parallel to compute $P_{\text{LLM}}(\hat{y}_t)$ and $P_{\text{SLM}}^-(\hat{y}_t)$ for each step. Since both models reside on the server, this incurs zero network latency.
    
    \item \textbf{Ratio-Based Decision:} The Cloud computes the ratio $\frac{P_{\text{LLM}}(\hat{y}_t)}{P_{\text{SLM}}^-(\hat{y}_t)}$ locally to determine acceptance.
    
    \item \textbf{Minimal Downlink (Steering Recovery):}
    \begin{itemize}
        \item \textbf{If accepted:} The Cloud sends a lightweight boolean confirmation (1 bit).
        \item \textbf{If rejected:} Instead of sending the full vocabulary distribution, the Cloud computes the non-private component of the steering objective: $h_{\text{cloud}} = h_{\text{LLM}} - \beta \cdot h_{\text{SLM}}^-$. To adhere to bandwidth constraints, we strictly truncate this vector to the top-$k$ tokens (e.g., $k=32$) with the highest values. The Local Edge receives this sparse vector, adds its private term $\beta \cdot h_{\text{SLM}}^+$, and performs the final resampling locally. This ensures the correction incorporates global reasoning while preserving local intent, with minimal data transfer.
    \end{itemize}
\end{enumerate}

By placing $\mathcal{M}_{\text{SLM}}^-$ on the cloud, we transform the verification into a purely server-side operation. This design reduces the communication payload from $\mathcal{O}(K \cdot V)$ (vocabulary size) to $\mathcal{O}(K)$ (token IDs), making real-time collaboration feasible even under limited bandwidth.


\begin{table*}[htbp]
    \centering
    \renewcommand{\arraystretch}{1.4} 
    \small 
    
    \begin{tabularx}{\textwidth}{l l l X}
        \toprule
        \textbf{ID} & \textbf{Task} & \textbf{Metric} & \textbf{ Sample} \\
        \midrule
        
        LaMP-1 & Citation Identification & Accuracy $\uparrow$ & \makecell[l]{\textbf{Input:} For author who wrote [content], which ref is related? \\ \textbf{Target:} [reference ID]} \\
        \cmidrule(lr){1-4}
        
        LaMP-2 & Movie Tagging & \makecell[l]{Accuracy $\uparrow$ \\ F1 $\uparrow$} & \makecell[l]{\textbf{Input:} For a movie described as [plot], select the best tag. \\ \textbf{Target:} [tag]} \\
        \cmidrule(lr){1-4}
        
        LaMP-3 & Product Rating & \makecell[l]{MAE $\downarrow$ \\ RMSE $\downarrow$} & \makecell[l]{\textbf{Input:} For a user reviewing [content], predict the score (1-5). \\ \textbf{Target:} [score]} \\
        \cmidrule(lr){1-4}
        
        LaMP-4 & News Headline Gen & \makecell[l]{ROUGE-1 $\uparrow$ \\ ROUGE-L $\uparrow$} & \makecell[l]{\textbf{Input:} Generate a headline for the article [content]. \\ \textbf{Target:} [headline]} \\
        \cmidrule(lr){1-4}
        
        LaMP-5 & Scholarly Title Gen & \makecell[l]{ROUGE-1 $\uparrow$ \\ ROUGE-L $\uparrow$} & \makecell[l]{\textbf{Input:} Generate a title for the abstract [content]. \\ \textbf{Target:} [title]} \\
        \cmidrule(lr){1-4}
        
        LaMP-7 & Tweet Paraphrasing & \makecell[l]{ROUGE-1 $\uparrow$ \\ ROUGE-L $\uparrow$} & \makecell[l]{\textbf{Input:} Paraphrase the tweet [content] to mimic user style. \\ \textbf{Target:} [paraphrased tweet]} \\
        \midrule
        
        LongLaMP-2 & Pers. Abstract Gen & \makecell[l]{ROUGE-1 $\uparrow$ \\ ROUGE-L $\uparrow$} & \makecell[l]{\textbf{Input:} Generate an abstract for title [title] using  [items]. \\ \textbf{Target:} [abstract]} \\
        \cmidrule(lr){1-4}
        
        LongLaMP-3 & Pers. Review Writing & \makecell[l]{ROUGE-1 $\uparrow$ \\ ROUGE-L $\uparrow$} & \makecell[l]{\textbf{Input:} Generate a review for [product] with  [score] and  [text]. \\ \textbf{Target:} [review body]} \\
        \cmidrule(lr){1-4}
        
        LongLaMP-4 & Pers. Topic Writing & \makecell[l]{ROUGE-1 $\uparrow$ \\ ROUGE-L $\uparrow$} & \makecell[l]{\textbf{Input:} Generate a Reddit post for [title] in the style of user [id]. \\ \textbf{Target:} [post content]} \\
        
        \bottomrule
    \end{tabularx}
    \caption{Overview of LaMP and LongLaMP tasks with metrics and samples.}
    \label{tab:lamp_longlamp_final}
\end{table*}

\definecolor{myblue}{RGB}{0, 0, 255}
\definecolor{myred}{RGB}{198, 40, 40}

\newcommand{\inputvar}{\textcolor{myred}{[INPUT]}}
\newcommand{\concat}{\textcolor{myblue}{\textbf{concat}}}
\newcommand{\addtitle}{\textcolor{myblue}{\textbf{add\_to\_paper\_title}}}

\begin{table*}[ht]
\centering
\small
\renewcommand{\arraystretch}{1.5} 
\begin{tabularx}{\textwidth}{@{}p{3.5cm}|X|X@{}}
\toprule
\textbf{Task} & \textbf{Per Profile Entry Prompt (PPEP)} & \textbf{Aggregated Input Prompt (AIP)} \\ \midrule

LaMP-1: Citation Identification & 
``$P_i$[title]'' & 
\addtitle(\concat([PPEP($P_1$), ..., PPEP($P_n$)], ``, and ''), \inputvar) \\ \hline

LaMP-2: Movie Tagging & 
the tag for the movie: ``$P_i$[description]'' is ``$P_i$[tag]'' & 
\concat([PPEP($P_1$), ..., PPEP($P_n$)], ``, and ''). \inputvar \\ \hline

LaMP-3: Product Rating & 
$P_i$[score] is the score for ``$P_i$[text]'' & 
\concat([PPEP($P_1$), ..., PPEP($P_n$)], ``, and ''). \inputvar \\ \hline

LaMP-4: News Headline Gen & 
``$P_i$[title]'' is the title for ``$P_i$[text]'' & 
\concat([PPEP($P_1$), ..., PPEP($P_n$)], ``, and ''). \inputvar \\ \hline

LaMP-5: Scholarly Title Gen & 
``$P_i$[title]'' is the title for ``$P_i$[abstract]'' & 
\concat([PPEP($P_1$), ..., PPEP($P_n$)], ``, and ''). Following the given patterns \inputvar \\ \hline

Lamp-7: Tweet Paraphrasing & 
``$P_i$[text]'' & 
\concat([PPEP($P_1$), ..., PPEP($P_n$)], ``, and '') are written by a person. Following the given patterns \inputvar \\ \hline

LongLaMP-2: Abstract  & 
Title: ``$P_i$[title]'' \newline Abstract: ``$P_i$[abstract]'' & 
Here are some examples of abstracts written by this author: \newline \concat([PPEP($P_1$), ..., PPEP($P_n$)], ``\textbackslash n\textbackslash n''). \newline Following these examples, \inputvar \\ \hline

LongLaMP-3: Review  & 
Product: ``$P_i$[description]'' \newline Summary: ``$P_i$[summary]'' \newline Review: ``$P_i$[reviewText]'' & 
Here are past reviews written by this user: \newline \concat([PPEP($P_1$), ..., PPEP($P_n$)], ``\textbackslash n\textbackslash n''). \newline Following their style, \inputvar \\ \hline

LongLaMP-4:  Writing & 
Summary: ``$P_i$[summary]'' \newline Content: ``$P_i$[content]'' & 
Here are past posts written by this author: \newline \concat([PPEP($P_1$), ..., PPEP($P_n$)], ``\textbackslash n\textbackslash n''). \newline Following their style, \inputvar \\ 

\bottomrule
\end{tabularx}
\caption{Prompts template used to augment the input of the LM with the user profile. \inputvar is the task's input as shown in Table \ref{tab:lamp_longlamp_final}. }
\label{tab:prompts}
\end{table*}

\begin{table*}[htbp]
\centering
\small
\renewcommand{\arraystretch}{1.5} 
\begin{tabularx}{\textwidth}{p{4cm} p{3.5cm} >{\RaggedRight\arraybackslash}X p{3cm}}
\toprule
\textbf{Dataset} & \textbf{Profile Format} & \textbf{Generated Input ($x_i$)} & \textbf{Generated Output ($y_i$)} \\ 
\midrule

LaMP-1: Personalized Citation Identification & 
title: [title] \newline abstract: [abstract] & 
Write an abstract for this title: [title] & 
[abstract] \\ \midrule

LaMP-2: Personalized Movie Tagging & 
description: [description] \newline tag: [tag] & 
Which tag does this movie relate to among the following tags? Just answer with the tag name without further explanation. tags: [sci-fi, based on a book, ... true story] description: [description] & 
[tag] \\ \midrule

LaMP-3: Personalized Product Rating & 
review: [review] \newline score: [score] & 
What is the score of the following review on a scale of 1 to 5? just answer with 1, 2, 3, 4, or 5 without further explanation. review: [review] & 
[score] \\ \midrule

LaMP-4: Personalized News Headline Generation & 
article: [article] \newline title: [title] & 
Generate a headline for the following article: [article] & 
[title] \\ \midrule

LaMP-5: Personalized Scholarly Title Generation & 
abstract: [abstract] \newline title: [title] & 
Generate a title for the following abstract of a paper: [abstract] & 
[title] \\ \midrule

LaMP-7: Personalized Tweet Paraphrasing & 
tweet: [tweet] & 
Complete the following tweet: [first part of the tweet] & 
[second part of the tweet] \\ \midrule

LongLaMP-2: Personalized Abstract Generation & 
title: [title] \newline abstract: [abstract] & 
Generate an abstract for the title ``[title]'' & 
[abstract] \\ \midrule

LongLaMP-3: Personalized Review Writing & 
rating: [overall] \newline desc: [description] \newline summary: [summary] \newline review: [reviewText] & 
Generate a review text written by a reviewer who has a given an overall rating of ``[overall]'' for a product with description ``[description]''. The summary of the review text is ``[summary]''. & 
[reviewText] \\ \midrule

LongLaMP-4: Personalized Topic Writing & 
summary: [summary] \newline content: [content] & 
Generate the content for a reddit post ``[summary]'' & 
[content] \\ 

\bottomrule
\end{tabularx}
\caption{Implementation of the input-output generation function \texttt{convert} for PEFT personalization on LaMP and LongLaMP benchmarks.}
\label{tab:peft_implementation}
\end{table*}

\section{Use of AI Assistants}

We used large language models only for language polishing and editing. The core method design, experimental execution, and initial manuscript drafting were completed independently by the authors.
\end{document}